# Learning mental states estimation through self-observation: a developmental synergy between intentions and beliefs representations in a deep-learning model of Theory of Mind

Francesca Bianco[a,b*], Silvia Rigato[b], Maria Laura Filippetti[b], Dimitri Ognibene[a,c,d]

[a]School of Computer Science and Electronic Engineering, University of Essex, Colchester, UK
[b]Centre for Brain Science, Department of Psychology, University of Essex, Colchester, UK
[c]Dipartimento di Psicologia, Università Milano-Bicocca, Milan, Italy
[d]Istituto di Scienze e Tecnologie della Cognizione, Consiglio Nazionale della Ricerche, Rome, Italy

*Corresponding author: francesca.bianco.lain@gmail.com

## Abstract

Theory of Mind (ToM), the ability to attribute beliefs, intentions, or mental states to others, is a crucial feature of human social interaction. In complex environments, where the human sensory system reaches its limits, behaviour is strongly driven by our beliefs about the state of the world around us. Accessing others' mental states, e.g., beliefs and intentions, allows for more effective social interactions in natural contexts. Yet, these variables are not directly observable, making understanding ToM a challenging quest of interest for different fields, including psychology, machine learning and robotics. In this paper, we contribute to this topic by showing a developmental synergy between learning to predict low-level mental states (e.g., intentions, goals) and attributing high-level ones (i.e., beliefs). Specifically, we assume that learning beliefs attribution can occur by observing one's own decision processes involving beliefs, e.g., in a partially observable environment. Using a simple feed-forward deep learning model, we show that, when learning to predict others' intentions and actions, more accurate predictions can be acquired earlier if beliefs attribution is learnt simultaneously. Furthermore, we show that the learning performance improves even when observed actors have a different embodiment than the observer and the gain is higher when observing beliefs-driven chunks of behaviour. We propose that our computational approach can inform the understanding of human social cognitive development and be relevant for the design of future adaptive social robots able to autonomously understand, assist, and learn from human interaction partners in novel natural environments and tasks.

# 1. Introduction

Humans' ability to represent and predict what others think or want from short interactions is crucial to our social lives (Baker et al., 2017). This ability to attribute mental states, such as intentions, desires and beliefs, to others is referred to as Theory of Mind (ToM), or mentalising (Frith & Frith, 2005). The importance of having a ToM for humans to successfully navigate the social world is widely recognised (e.g., for collaboration, communication, imitation, teaching, deceiving, and persuading (Devaine et al., 2014, Frith & Frith, 1999; Kovács et al., 2010; Rakoczy, 2017; Tomasello et al., 2005)). Furthermore, its critical role in social cognition has also been evidenced by studies assessing individuals with neurodevelopmental disorders characterised by social deficits (e.g., autism spectrum disorder (ASD)) who tend to display atypical ToM ability (e.g., Baron-Cohen et al., 1985; Murray et al., 2017; Schneider et al., 2013). This capacity is not limited to the ability to process others' behaviours (Frith & Frith, 1999). Knowing that other individuals have mental states which may differ or contrast with our own, that may be inaccurate with respect to the state of the world, and that drive their behaviour, is fundamental for efficient and fluid human social interaction (Frith & Frith, 1999; Rakoczy, 2017). Hamlin et al. (2013) have further distinguished ToM ability into "full ToM" (or high-level ToM), which assumes second-order beliefs (i.e., the realisation that it is possible to hold a false belief about someone else's belief) vs "mid ToM" (or low-level ToM), which assumes second-order goals (i.e., the realisation that the goal of one agent depends on the goal of another agent) but not second-order beliefs.

Given the implications of ToM in social cognition and human-human interactions, it has been previously suggested that equipping robots with a ToM would also improve human-robot interactions (HRI). Indeed, with robots increasingly becoming part of the society, achieving more natural and successful HRI is important, and providing robotic architectures with a ToM is considered one of the "Grand Challenges of Science Robotics" (Yang et al., 2018, p. 9). In the last few decades, the fields of artificial intelligence and robotics have greatly advanced, resulting in the development of increasingly sophisticated virtual and physical intelligent agents with complex abilities and behaviours (e.g., Bhat et al., 2016; Görür et al., 2017; Hoffmann et al., 2017; Milliez et al., 2014). AI and robotic agents' increasingly humanoid features and human-inspired complex behaviours have enhanced humans' positive attitude towards them (Hegel et al., 2008; Rossi et al., 2023; Thellman et al., 2022). Nonetheless, the integration of AI and robots among humans is still far from optimal. This may be due to two main reasons. On the one hand, AI's still limited social capabilities continue to hinder humans' acceptance of AI in their daily lives and of robots as social companions (Abubshait & Wiese, 2017). Considerable steps were accomplished in this direction with the recent advances of AI in the linguistic domain, where an improvement in ToM-related tasks was seen with Large Language Models like chatGPT (Kosinski, 2023; Soubki et al., 2024). However, their performance is still far below human level and it further decreases when physical aspects are taken into account (e.g., lack of embodiment within an action-oriented environment), limiting their application in robots and embodied settings (Ma et al., 2023; Sap et al., 2022; Strachan et al., 2023). On the other hand, humans have been often seen as a source of complexity, disturbance, and unpredictability that could affect autonomous agents' performance (e.g., Hiatt et al., 2011; Koay et al., 2007; Sisbot et al., 2007), thus limiting AI and robotic agents' application. Equipping robots with a ToM would aid both these issues. Indeed, an artificial ToM would endow robots and intelligent agents with increased social capabilities, by being able to learn, represent, and reason about mental states and appropriately react to them. In addition, it would allow them to understand mental states-driven human behaviour, thus decreasing humans' unpredictability and disturbance, resulting in more fluid and efficient HRI.

While humans provide the best example of effective ToM (Tomasello et al., 2005), the literature suggests that we still know relatively little about ToM. Although decades of research in various disciplines, from psychology and neuroscience to artificial intelligence and robotics (Baron-Cohen et al., 1985; Devin & Alami, 2016; Frith & Frith, 2006; Rabinowitz et al., 2018), have been dedicated to human ToM ability, key features of this cognitive skill are yet to be fully described. There are currently several debates that surround ToM, including (1) when does ToM emerge; (2) which biological and computational mechanisms underlie ToM and its development; (3) which factors (e.g., self-other similarity) are implicated in ToM ability and its development (Bianco & Ognibene, 2019, 2020; Butterfill & Apperly, 2013; Kampis et al., 2015; Skerry et al., 2013), and (4) whether machine ToM is achievable (Yang et al., 2018). While we contribute to all these debates with this paper, our primary emphasis lies on the computational mechanisms that underlie ToM and its development. In particular, our specific objective is to propose and evaluate a novel computational framework able to support the early learning of mental state attribution in humans, thereby contributing valuable insights for the design of social robotic architectures. Below, we provide a brief overview of these long standing debates, highlighting the pertinence of our innovative computational framework to the majority of these discussions.

## 1.1 Related Work

**1.1.1 Early vs late ToM emergence.** While it is now widely accepted that children older than 4 years of age show evidence of ToM ability (e.g., Wellman et al., 2001), whether this cognitive ability can be observed at younger ages is yet to be confirmed (Apperly & Butterfill, 2009; Ruffman & Perner, 2005; Sodian & Kristen, 2016). It has been previously indicated that implicit (or spontaneous) ToM competencies may exist from early infancy, and that only later during development, once children overcome verbal and inhibitory demands, this ability becomes explicit (e.g., Sodian et al., 2020). However, this remains debated (Sodian et al., 2020; van Overwalle & Vandekerckhove, 2013). Findings in support of early ToM ability in infants come from behavioural, neuroimaging, and computational studies (see *Supplemental Table 1* for further details on these studies and measures). For example, a study by Southgate and Vernetti (2014) investigated early ToM ability in 6-month-old infants by assessing their brain activity through electroencephalography (EEG) in response to an agent's varying belief states and related actions. The results from this study suggest that 6-month-old infants are able to both create and update representations of another person's beliefs, even when these are incongruent to their own beliefs, and that such representations guide their predictions of that person's future behaviour. From a computational perspective, Hamlin et al. (2013) evaluated infants' understanding of others' beliefs by testing which of the following strategies would drive 10-month-old infants' judgements in a social evaluation task: (1) the analysis of the mental states at play (i.e., the protagonist's preference for an object and the helper's knowledge of such preference), (2) the assignment of the same knowledge about the protagonist's mental states to both the protagonist and the helper, or (3) the reliance on low-level cues in the scene (e.g., valence-matching hypothesis (Scarf et al., 2012)). To do so, the authors compared a Bayesian computational ToM model with simpler models that rely on cue-based and goal-attribution accounts (instead of full ToM), which were used in combination with the more classical behavioural measure of ToM (looking behaviour). The authors concluded that the Bayesian computational ToM model best described infant looking behaviour in their task, providing evidence of ToM processing from an early age rather than non-ToM computations. By investigating the developmental trajectory and learning of mental state attribution in a computational model, this paper contributes to the debate surrounding ToM emergence by providing a stance supported by additional computational findings.

**1.1.2 Mechanisms underlying ToM.** An important point of discussion concerns the non-ToM-specific computational processes that may underlie infants' observed behaviour in ToM studies. Specifically, three main psychological intention recognition theories, i.e., action-effect association (Csibra & Gergely, 2007), simulation (Rizzolatti et al., 2001) and teleological (Csibra & Gergely, 2007) accounts, have been proposed in an attempt to explain how infants' performance in false-belief tasks may result from intention recognition abilities that are considered different to full or higher-level ToM (see also *Supplemental Table 2*). These theories differ from ToM in that the computational processes they describe do not fully explain humans' ability to infer mental states of other agents (ToM components). Indeed, while they may allow the understanding and prediction of others' intentions (or goals) from their observable behaviour (Gergely & Csibra, 1997; Monroy et al., 2017; Skerry et al., 2013; Southgate et al., 2008; Woodward, 1998), ToM pinpoints to the unobservable, internal causal structures underlying behaviour, i.e., mental states, which go beyond action observation (Baird & Astington, 2005).

First, the action-effect association theory states that goals are simply inferred by learnt associations between an observed action and the effects that such action has produced (Csibra & Gergely, 2007). Evidence for this theory comes from Woodward's (1998) study with 6-month-olds, who reportedly showed the ability to attribute goal inference directed by an association created between previously observed actions and the effects of that action.
Second, the simulation theory proposes that actions are understood when the observer directly matches, or mirrors, the observed action onto their own motor system (Rizzolatti et al., 2001). Southgate et al. (2009)'s neuroimaging study provides an example of 9-month-old infants engaging in simulation for others' action prediction. In this study, using EEG during a reaching paradigm, the authors reported attenuation of sensorimotor alpha band activity (i.e., neural indicator of action prediction) in 9-month-old infants both when reaching for an object themselves, when observing an agent reaching for an object and before the agent commenced moving. The authors therefore suggested the involvement of simulation processes in infants' prediction of others' behaviour. Third, the teleological theory proposes that outcomes of actions can be recognised as their goals only if they are performed efficiently (Csibra & Gergely, 2007). Supporting evidence includes Southgate et al.'s (2008) study in which 6-8-month-old infants were found to attribute goal-directedness in a reaching task also when biomechanically impossible actions were completed to reach the target, as long as these actions were efficient. In concordance with the teleological account, this study suggests that infants assign goal-directedness to actions, as long as such action is efficient.

These intention recognition theories have been previously used to rule out full ToM in infants and children undertaking false-belief tasks, in an attempt to explain infants' success in such tasks with simpler, non-ToM interpretations. For example, as opposed to a ToM interpretation, the teleological theory was used to interpret false-belief task results in 18-month-old infants (Priewasser et al., 2018), as well as the association (De Bruin & Newen, 2012) and simulation (Asakura & Inui, 2016) theories in 3- to 6-year-old children. However, it is still unclear whether these theories can be accounted as computational processes underlying ToM or be considered as precursors of ToM. Some authors (e.g., Asakura & Inui, 2016; Keysers & Gazzola, 2007) have previously supported the stance of psychological intention recognition theories as precursors of ToM. Specifically, Asakura and Inui (2016) proposed an integration in the same framework of the simulation and teleological theories for false-belief reasoning. Keysers and Gazzola (2007) brought forward the idea of a continuum between simulation and ToM. See also *Supplemental Table 3* for a summary of psychology studies describing these intentions recognition mechanisms as precursors of ToM. By drawing inspiration from these psychology theories and applying them to a novel human-inspired computational framework of ToM, this paper contributes to the open question of the mechanisms underlying ToM.

**1.1.3 Computational models of ToM and Machine ToM.** Previous seminal work presented compelling artificial architectures for social agents able to predict their partner's intentions or computational models for the study of the implementation of ToM functions (e.g., Baker et al., 2017; Devin & Alami, 2016; Giese & Rizzolatti, 2015; Görür et al., 2017; He et al., 2016; Rabinowitz et al., 2018; Raileanu et al., 2018; Winfield, 2018; Nguyen & Gonzalez, 2022; Chen et al., 2021; see Langley et al., 2022 for a recent review of ToM computation models). For example, Devin and Alami (2016) integrated a ToM model in their robot control architecture for successful human-robot shared-plans performance. This architecture adopts a symbolic approach and assumes a readily available hand-crafted list of possible goals and actions with their models, as well as full knowledge of the partner's awareness of all the effects of such actions. The state of the plan in the partner's mental state can then be estimated by observing the partner's actions, and be updated following unexpected or failed actions. Görür et al. (2017) used a Hidden Markov Model (HMM) to estimate actions performed by interacting agents in a collaborative task and incorporate human emotional states. Similarly to the previous paper, this system relies on a readily available hand-crafted set of actions and states. Estimation of the partner's action state is possible by tracking their actions, with the assumption that all observations are discrete and emotional states, which are explicitly provided to the robot in the form of reactions. Winfield (2018) implemented a simulative computational model for ToM where robots are provided with an internal model of themselves and their environment, including other agents. This model could test (i.e., simulate) next possible actions and anticipate likely consequences, both of the robot itself and others. Baker et al. (2017) presented a compelling Bayesian computational model of ToM (BToM), which was found to accurately infer mental state judgements of human participants. This computational model formalises ToM as a Bayesian inference about unobserved mental states (beliefs, desires, percepts) of a Partially Observable Markov Decision Process (POMDP) agent, based on observed actions. While providing important experimental results on human processing, this work makes strong assumptions for a practical implementation in a robot. For example, the observer is aware of the specific POMDP model that the observed agent uses; in addition, inverting this model for inference can be computationally heavy. Raileanu et al. (2018) and He et al. (2016) used reinforcement learning for equipping artificial agents with the ability to learn to collaborate with others based on their intentions and focused on fully observable settings. Briefly, Raileanu et al. (2018) introduced the Self Other-Modeling (SOM) approach to solve multi-agent adversarial and cooperative tasks, driven by a reward function based on intentions of both agents in the task. Similarly to other simulation models (e.g., Demiris, 2007; Ognibene & Demiris, 2013), the SOM approach allowed the agent to use its own policy to predict the other agent's actions and underlying intentions in an online manner. During the game, the agent inferred the other agent's hidden goal by directly optimising through gradient descent over the other goal (represented as a Gumbel-Softmax distribution) using its own action model to maximise the likelihood of others' observed actions. He et al. (2016) instead focused on jointly learning an observer's policy and the opponents' behaviour and mode/goal (defensive vs offensive) using a deep Q-Network, specifically applied to games like simulated soccer. The authors of this paper resorted to multitasking to explicitly model other agents' actions and strategy, using the following two supervisory signals, i.e., others' actions in the current state and their goal. Chen et al., (2021) presented a visual-based ToM model implemented as a convolutional neural network-based system observing simple scenes. This model, after watching the actor (i.e., observed acting agent) act in various situations, is capable of predicting the actor robot's behaviour in novel, previously unseen situations.

While these studies and others (see *Supplemental Table 4*) provided compelling computational models of ToM and architectures for social robots, they mainly focused on the final performance of intention prediction, often disregarding the learning of explicit representations of others' beliefs for intention and behaviour prediction.

A recent study performed by Rabinowitz et al. (2018) partly addressed the above by exploring the learning of explicit belief representations (thus ToM) by a meta-cognitive agent for belief-based behaviours prediction. Specifically, the authors trained a ToM agent to observe only others' overt behaviour, and used it to predict, for a given query state, the others' policy, consumptions, successor representations, and reported beliefs (i.e., what the agent would say about the world if it were asked). To achieve this, the authors constructed a supervised dataset of behaviours of reinforcement learning-based agents (Jaderberg et al., 2017), with limited perception but endowed with explicit belief states, dwelling in a 11x11 gridworld. A similar setting was assumed by Nguyen & Gonzalez (2022) but adopting the Instance-Based Learning Theory framework (Gonzalez, Lerch, & Lebiere, 2003) and ACT-R cognitive architecture (Anderson & Lebier, 2013).

While Rabinowitz et al. (2018) assumed a rich communication channel to access other's beliefs during learning, they also brought forward the idea of accessing one's own latent mental states during action execution through meta-cognition. However, the authors deemed this approach to be limited and likely not to scale to real-world situations. In particular, they suggested that the self-generated supervisory signal needed to explicitly learn beliefs, as accessed through one's own mental states made available by meta-cognition, may be too biased. Sample efficiency of this approach was not assessed, and the (developmental) learning trajectories of their belief-based behaviours prediction models were not compared with classical models of intention prediction not comprising belief learning. This comparison is particularly important here considering that deep convolutional networks, adopted in that study for their high prediction accuracy, are known to be data hungry. Therefore, lack of data can be associated with misinterpreting others' actions due to the high variability of social interactions and agents' behaviours, while too much data is associated with high ecological cost. In this paper, we assess the constraints of learning to explicitly represent beliefs on others' behaviour prediction performance, building upon the framework introduced by Rabinowitz et al. (2018). We delve into the challenges posed by restricted access to and insufficient quantities of belief data. This analysis enables us to contribute to the ongoing debate surrounding the feasibility of achieving Machine ToM, while considering the parallel between data availability and experience acquisition during childhood.

## 1.2 Our approach

In the present paper, using an adaptation of the deep-learning model of ToM presented in Rabinowitz et al. (2018) and exploiting its multi-task learning, we aim to contribute with a computational modelling approach to the open debates surrounding ToM introduced above, i.e., early vs late ToM emergence, the mechanisms and factors underlying ToM, and inform computational models of (machine) ToM.

**1.2.1 Early vs late ToM emergence.** We explore the developmental trajectory of explicitly representing others' beliefs by simulating different levels of experience (training data size). Through ablation, we further delve into the nature of the impact that acquiring representations of others' beliefs has on predicting intentions, thereby emphasising the ecological significance of early acquisition of full ToM for predicting behaviour during social interactions. Rabinowitz's (2018) approach is particularly fit to provide a baseline on the combined impact of experience and belief estimation as their system learns to infer others' beliefs and intentions from scratch, making few assumptions about the processes driving observed behaviours.

**1.2.2 Mechanisms underlying ToM.** We introduce the "like-them" assumption based on the association mechanism for ToM (see *Figure 1*). In more detail, we borrow from the "like me" theory of social cognition (Meltzoff, 2007a, 2007b) the assumption of a shared abstract structural framework that facilitates knowledge transfer between self and others' representations. However, to understand observed others' behaviours (and underlying mental states), our "like-them" variation does not rely on physically shared representations as those that would be embodied in the observer's control and planning systems (simulation mechanism). In contrast, this process is inverted. Instead of generating trajectories on the fly using the observer's control and planning system, the "like them" assumption crucially posits that, through a mechanism of self-observation, one's own mental states during task execution are presumed to be similar to future putative mental states and actions of other agents ("like them") engaged in analogous or similar tasks. Therefore, own (mental) states and actions provide a pair of supervisory signals and observed samples to train a predictor of others' mental states. This "like them" assumption can also be considered in line with the associative hypothesis, which has been described by James (1890) as a mechanism linking action-effect representations through bidirectional associations. Our approach builds upon previous research (e.g., Hommel et al., 2001, as described in Csibra & Gergely, 2007) which expanded upon this associative hypothesis for the interpretation of others' intentions, and further extends it to beliefs. Specifically, our 'like-them' approach proposes that associations between behaviours (actions) and underlying intentions (effect) develop from an early age and can be used later in life as a means to infer and predict others' intentions and behaviours. The associative account has also been previously related to the understanding of others' behaviours by proposals of its involvement in mirror neurons development (e.g., Heyes, 2010), as well as in sensorimotor matching for imitation (Decety & Chaminade, 2003). Our work differs from these accounts as it focuses on predicting others' beliefs. It relies on associations learnt through self-experience between own belief representations and consequent beliefs-driven behaviours to improve prediction performance of others' intentions. We also investigate whether our computational model could be extended to interpret behaviours of actors with different cognitive and physical abilities to the self. This will shed light on factors important for engaging in ToM, such as whether self-other similarity between the observer and observed agent is a requirement for successful ToM, and for its development.

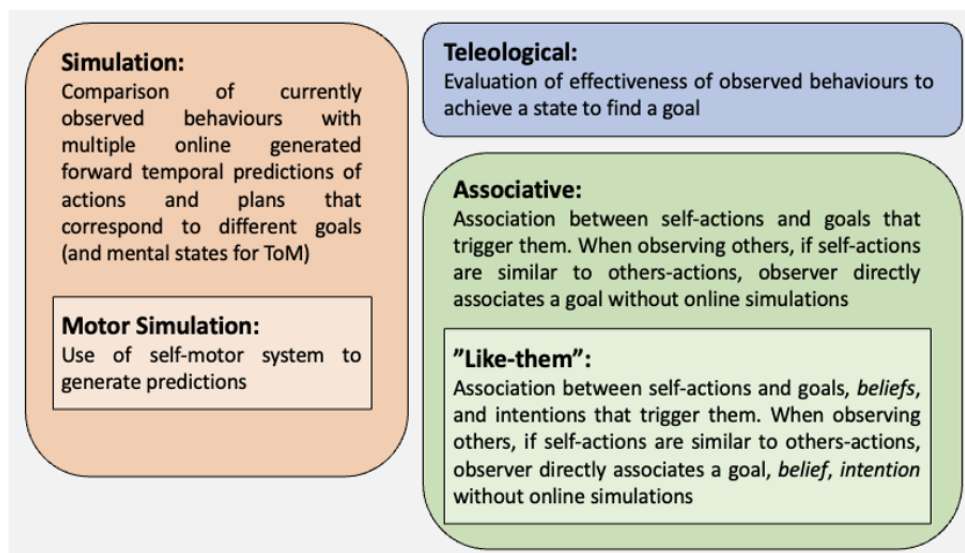

**Figure 1.** "Like-them" assumption within the context of other psychological intention recognition theories. The "Like-them" assumption can be considered as a component of the association mechanism and separate from the simulation and teleological mechanisms for Theory of Mind.

**1.2.3 Machine ToM.** Using a simple feed-forward deep learning model, we conducted a series of experiments to explore the suitability of employing this computational architecture for attaining machine ToM. Starting from human-inspired computational modelling, we interpreted the results from a robotics perspective to inform the development of artificial architectures and social agents equipped with adaptive ToM.

Overall, our results show a possible synergy between the prediction of others' intentional behaviour and the explicit representation of others' beliefs from an early age, likely resulting from a regularisation effect of multitask learning (Crawshaw, 2020; Ruder, 2017). We show this effect without forcing an explicit interaction between the two processes, which may have resulted in an even higher improvement in behaviour prediction performance. Furthermore, our results show that self-observation (of own behaviours and processes) can be effectively used as a training signal ("like-them" assumption based on association mechanisms for ToM). The positive impact of learning beliefs attribution towards intention prediction is stronger in conditions of partial observability, e.g., when the observed actor does not perceive or even know the location of its target. This last condition closely resembles the false-belief scenario typically employed in the study of ToM. Finally, our results indicate that this positive impact can be generalised to actors with different cognitive and physical abilities, suggesting that self-other similarity is not a strict requirement for successful ToM. Although a slight decrease in behaviour prediction performance is seen when the observer and actor are increasingly different. From a computational modelling perspective, our work indicates that developing the capability of learning to explicitly represent beliefs through a shared abstract representational framework (a) poses architectural demands that are simpler than previously believed, (b) contributes in speeding-up the acquisition of socio-cognitive prediction skills, (c) strongly improves the interpretation of beliefs-driven behaviours, and (d) increases the generalisation ability to predict behaviours of others acting in different environments. We thus highlight a usable approach for adaptive and social robots with increasing ToM skills.

## 2. Methods

### 2.1 Architecture

The architecture in this study consists of a Social Perception System (SPS) and an Agent Control System (ACS) (see *Figure 2*). The SPS interprets and allows for predictions of the observed actor's behaviour (i.e., next actions, goals and beliefs). The ACS defines the actor's behaviour based on tasks interpreted in the SPS. The quantities emerging from the control system, i.e., actions, intentions and beliefs, are also used to fully train the SPS; in contrast, others' behaviours are only used for partial training, as others' beliefs cannot be observed.

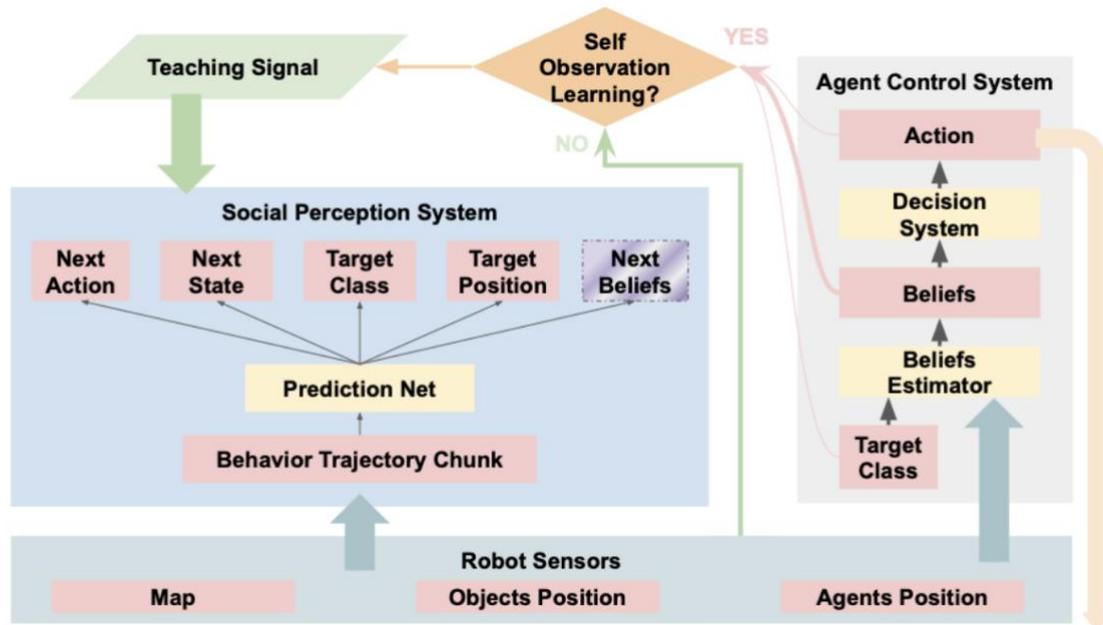

**Figure 2.** Description of the architecture utilised in the reported studies. The architecture consists of a Social Perception System (SPS) and an Agent Control System (ACS). The SPS takes a behaviour trajectory chunk from the incoming sensory information and predicts the observed actor's behaviour (i.e., next actions and states, goals and beliefs) through a shared prediction net torso which subsequently separates into prediction heads. For the Beliefs architecture, the Beliefs prediction head (in purple) is also considered. The ACS defines the agent's behaviour. Under the "like-them" assumption, own (mental) states and actions generated in ACS together with sensory observation provide a pair of supervisory signals and observed samples to train a SPS, predictor of others' mental states.

2.1.1 Social Perception System (SPS): The SPS's goal is to make predictions about the observed actor's future behaviour, with a specific interest on the actor's target position. Two types of SPS were trained, the Beliefs SPS (the ToM agent) and the NoBeliefs SPS (the classical agent), i.e., an ablation of the Beliefs SPS deprived of the explicit belief prediction. They both predict an actor's target position, next action, and next resulting state; however, only the latter has the additional ability to predict beliefs as well.

2.1.1.1 Input sensing and routing: The SPS can observe themselves or others. The input vector for the system can then be provided by a common reference frame to represent either the observer's self-localisation state (during self-observation learning), or the physical state of the other actor (both for learning and prediction of other actor's behaviour). Several architectures (e.g., Demiris & Johnson, 2003; Wolpert et al., 2003) have studied the problem of how to switch between the processing of one's own and others' state and how to perceive or infer others' physical states. Differently from motor-based simulation models (which also rely on switching the input for their social perception system), in our case, perceiving others' state will not interfere with behaviour execution. Conversely, training the SPS through self-observation during execution of a task may affect prediction of others' mental states.

2.1.1.2 Input encoding and pre-processing: The input consists of a number (max 5 in the reported experiment) of past steps of an actor's trajectory on a single grid map and the map itself. In particular, observed actions-states pairs are combined through a spatialisation concatenation operation, whereby actions are tiled over space into a tensor and concatenated to form a single tensor of shape (11 x 11 x 20). While 11 x 11 represents the size of the gridworld environments, 20 vectors are provided as inputs consisting of information regarding (a) actions (9 possible actions in experiments, thus 9 vectors); (b) objects coordinates, including the position of the actual target and that of distractors (4 objects, thus 4 vectors, one for each object class); (c) actor's position in past steps (5 past steps, thus 5 vectors); (d) 1 feature plane for the walls in environments; and (e) 1 vector for the actor's current position.

*2.1.1.3 Prediction Net:* Following spatialisation concatenation operation, tensors are passed through a deep learning architecture. The architectures utilised to conduct the reported experiments are formed of a shared prediction net trunk (Crawshaw, 2020) and subsequently of separate prediction heads.

• The trunk is implemented as a 2-layer ResNet with 32 channels, leaky ReLU nonlinearities, and batch-norm. The heads have all a similar architecture but differ in terms of output size and format.

• Target Location Prediction Head. The output from the trunk is inputted into a 1-layer Convnet with 32 channels and leaky RELU, another 1-layer Convnet with 16 channels and leaky RELU, followed separately by (a) a fully connected layer to 121-dim logits (11 x 11 gridworld) and (b) another 1-layer Convnet with 4 channels to 1. These are then summed, thus following a residual network approach.

• Next Action Prediction Head. The output from the trunk is inputted into a 1-layer Convnet with 32 channels and leaky RELU, followed by average pooling, and 2 fully connected layers to 9-dimensions (9 possible actions).

• Next State Prediction Head. The output from the trunk is inputted into a 1-layer Convnet with 32 channels and leaky RELU, another 1-layer Convnet with 16 channels and leaky RELU, followed separately by (a) a fully connected layer to 121-dim logits (11 x 11 gridworld) and (b) another 1-layer Convnet with 4 channels to 1. These are then summed.

• Beliefs Prediction Head. The output from the trunk is inputted into a 1-layer Convnet with 32 channels and leaky RELU, another 1-layer Convnet with 16 channels and leaky RELU, followed separately by (a) a fully connected layer to 121-dim logits (11 x 11 gridworld) and (b) another 1-layer Convnet with 4 channels to 1. These are then summed.

2.1.2 Agent Control System (ACS): The ACS defines the agent's behaviour. Under the "like-them" assumption, own (mental) states and actions generated in ACS together with sensory observation provide a pair of supervisory signals and observed samples to train a SPS, predictor of others' mental states. The actor has full information about its own position in the environment and partial observability over the target position in the environment, i.e., they could see it only when it was in their 5x5 field of view. To account for the resulting uncertainty and related information-gathering behaviours (Friston et al., 2015), the actors' trajectories were generated using the POMDP planner based on Monte Carlo tree search proposed by (Ognibene et al., 2019b) which extends (Silver & Veness, 2010) and integrates a Bayesian filter that explicitly represents the actor's beliefs about the state of the task, i.e., the probability distribution on the target position. Silver & Veness (2010) shows that the actor's belief and related uncertainty reduction allows efficient sampling. In the current paper, the actor's belief in the starting particle can be used as a training objective for the belief prediction head. Note that while the beliefs design of the observed actor are necessarily determined by its task (i.e., represent the target position estimation), this does not affect the generality of the observer's performance because it is blind to the task and belief design assumptions.

2.2 Environments

For all experiments in this study, the environments consisted of 11x11 gridworld maps, which varied in the location of walls, columns, and free cells to move around the map (see *Figure 3* for a visualisation of example random maps and connected behaviours). To assess the impact of dataset size on the proposed approach, we created multiple training datasets comprising 5, 10, 15, 20, 25, 30, 60, 120, and 300 maps, together with a 10 maps testing set. The environments enabled a common action space (north, east, south, west, northeast, northwest, southeast, southwest, stay) with deterministic results. During each trial, both in training and testing, the target (in yellow in Figure 3 below), the distractors (green), and the actor (blue) are randomly positioned on free cells. The actor's previous positions (if present) are also provided (pink) in the gridworld map.

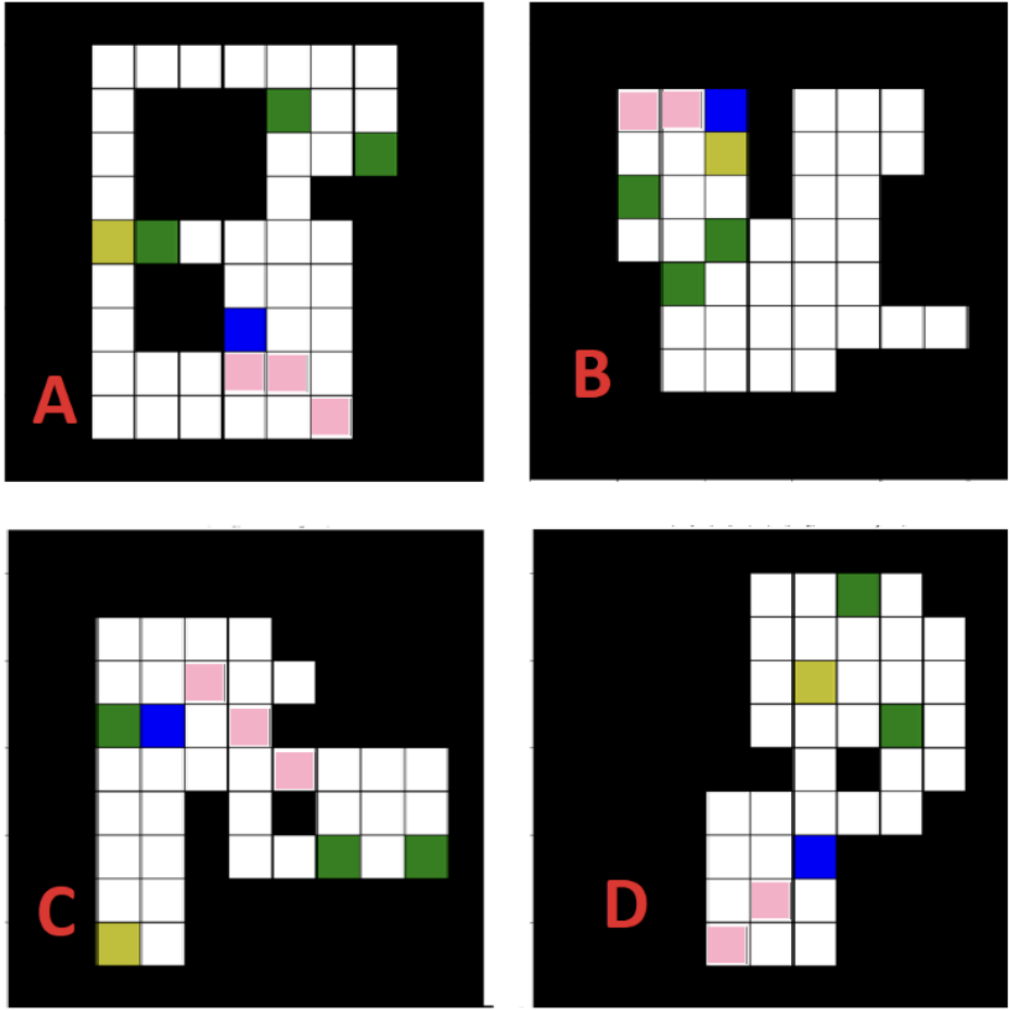

**Figure 3.** Visualisation of example 11x11 grid world maps. The maps varied in the location of walls, columns and free cells to move around the map. Colour code - Black: walls; White: empty cells; Yellow: target; Green: distractor objects; Blue: current actor's position; Pink: actor's past positions.

2.3 Trajectory Dataset Generation

A total of 30 trajectories were generated per gridworld map by randomly selecting for each trajectory the initial locations of both the actor and target. For example, when considering the dataset with 60 gridworld maps, a total of 1800 actors' behaviours were created, while 9000 total behaviours were created for the dataset with 300 maps. When considering all generated actors' behaviours from all gridworld maps datasets, the actor took, on average, approximately 3 steps ($\sigma^2$ = ~11) to reach its target and complete the task. Furthermore, on average, the target was not visible to the actor for approximately 3 steps ($\sigma^2$ = ~12), while the actor took, on average, approximately 3 steps ($\sigma^2$ = ~ 10) to reach its target once it became visible. At training and test time, 3 distractor objects, identical to the target, were positioned at different empty locations on the map. The positioning of the distractor objects was random, unless otherwise specified. This enabled a wider environment and contextualised behavioural diversity with no additional computational cost.

2.4 Training

All architectures (both Beliefs and NoBeliefs) were trained with the Adam optimiser,

with varying learning rates (we tested 6 levels from 0.00015 to 0.001), using batches of size 32. A learning rate scheduler with the following parameters was used in all experiments: milestones = [30, 60, 80, 160], gamma = 0.5. The Cross Entropy loss function was utilised for all heads training, except for the Belief head, for which the Kullback–Leibler divergence loss function was used instead. This choice was driven by the fact that the Action, State and Target prediction heads all required one-hot encoding to identify one single position in the map, whereas a distribution of probabilities over each map location was needed for the Belief prediction head. Losses calculated for all prediction heads were linearly combined with linear factors extracted during a preliminary tuning phase. A preliminary tuning of L1 and L2 regularisation was performed using the L1 and L2 factor search; the final values of the L1 and L2 parameters used in all experiments were 0.005 and 0.001, respectively. In addition, early stopping was also integrated during training as a means to prevent overfitting.

## 3. Results and Discussion

3.1 Developmental synergy between the prediction of others' intentional behaviour and the explicit representation of others' beliefs

*Figure 4* shows the performance of the NoBeliefs and Beliefs architectures evaluated over the whole test set with an increasing number of maps made available during training. Overall, the Beliefs architecture has better performance compared to the NoBeliefs architecture, achieving best accuracy of 72.43% (Beliefs, 300 maps). The gain becomes statistically significant (p value < 0.01) after the agent could observe the task over > 20 training maps (600 task executions). The worst observed performance is 59.26% by the Beliefs architecture when training with 5 maps (150 observed trajectories or task executions). Given that performance variance across trained nets is high (~19%) with this limited dataset, these low performances are not statistically significant. However, as seen in the learning trajectory, with a more reasonable dataset size, beliefs' prediction contributes to achieve higher performance. The maximum performance gain of the Belief architecture over the NoBelief architecture is 1.89% (69.45% vs 67.57%, respectively, p value < 0.001) when the agent was trained with 25 maps (750 task executions). An interesting trend to note here is that the positive contribution of learning beliefs' representations is stronger in the middle of the learning trajectory. This behaviour suggests a regularisation effect, which becomes important when the architecture stops memorising beliefs samples, and in turn is redundant with a high number of samples. Interestingly this positive contribution is present when analysing the complete dataset, which contains several conditions where the actor's target would be visible making the prediction of the target and that of the beliefs identical and redundant.

From a ToM point of view, these results indicate that beliefs processing may be an ability that develops during infancy in humans and can be used to aid the interpretation of others' behaviours from such an early age, supporting early ToM emergence as suggested in psychology literature (e.g., Southgate and Vernetti, 2014; Hamlin et al., 2013; see also Supplemental Table 1). Nevertheless, the results also indicate that beliefs processing starts playing an increasingly important role for understanding others' behaviours with increasing experience until reaching a possible plateau of maximum impact. After this, beliefs again gradually become less useful for predicting others' behaviour, while still representing a good source of information. This is also in concordance with previous literature suggesting ToM's further development during adolescence (e.g., Meinhardt-Injac et al., 2020) and middle adulthood (e.g., Spenser et al., 2020), i.e., increasing experience, and then reaching a plateau or even a decrease in ToM ability in late adulthood (e.g., Pearlman-Avnion et al., 2018). Overall, these findings, which indicate a beneficial role of beliefs in predicting others' intentions from infancy and throughout development, seem to provide a computational foundation for previous psychology research on ToM emergence and development. In the next experiments we explore the impact of beliefs-driven behaviour on intention prediction.

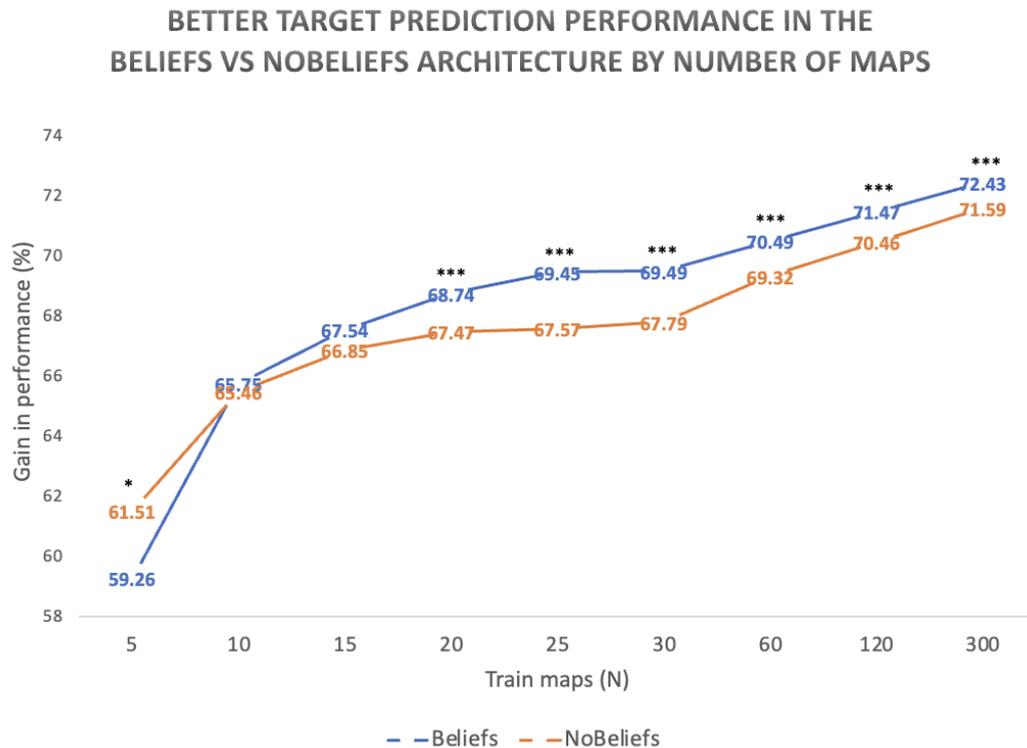

**Figure 4.** Increase in performance by the Beliefs vs NoBeliefs architectures with an increasing number of training maps. Average target prediction accuracy at each number of map; ***: significant at the .001 level (2-tailed); *: significant at the .05 level (2-tailed).

3.2 Maximal synergy with hidden target

Beliefs representations determine behavioural chunks when the target object is not yet visible to the actor and its position is not deterministically known. In such conditions, as shown in *Figure 5*, the performance difference between the Beliefs and NoBeliefs architectures peaks. A maximum difference in performance of ~7% (p value < 0.001) is provided by the Beliefs (58.17%) over the NoBeliefs (51.11%) architecture when trained with 15 maps (or 450 task executions), thus showing a relative gain of 14%. We must note that this environmental setting is more challenging for the observing agents, as we see the worst performance at 37.44% and the best at 63.11%. We indeed reasoned that unknown target position would produce wandering behaviours in the observed actor, which are not informative with respect to the actual target position. This being said, as performance was substantially above chance (25%), we next explored the dynamics by which the SPS could extract information from the presence of non-target objects, i.e., distractors, whose configuration could significantly affect intention prediction performance.

From a ToM perspective, these results suggest that being able to process beliefs may be mostly informative towards behaviour prediction when observing individuals engaging in exploratory behaviours. In exploratory conditions, individuals indeed rely more strongly on their beliefs (Knox et al., 2012); thus, if an observer is aware of them (i.e., is endowed with a ToM), a substantial gain in performance may be obtained. False-belief tasks can be considered the golden example of this condition, as they require beliefs processing for behaviour prediction of agents with partial knowledge about the environment. Indeed, observers need to predict and interpret the behaviour of observed agents that have false beliefs about the state of the world due to partial knowledge about the environment (Wimmer and Perner, 1983), e.g., the observed agent (falsely) believes an object to be in a box as they did not see that the object was previously removed from the box by another agent. Our results show a developmental trend with beliefs processing being impactful towards prediction of exploratory behaviours with minimal experience, reaching highest impact at a medium level of experience, and showing decreased impact with high levels of experience. Therefore, these results seemingly support previous literature - and in particular those studies that showed infants' ability to successfully complete false belief tasks (e.g., Southgate and Vernetti, 2014; Hamlin et al., 2013; see also Supplemental Table 1) - in favour of an existing ToM capability in infancy. Furthermore, our findings again support ToM further development with increasing experience during adolescence (e.g., Meinhardt-Injac et al., 2020) and middle adulthood (e.g., Spenser et al., 2020) until reaching a plateau and / or a decrease in ToM ability in late adulthood (e.g., Pearlman-Avnion et al., 2018).

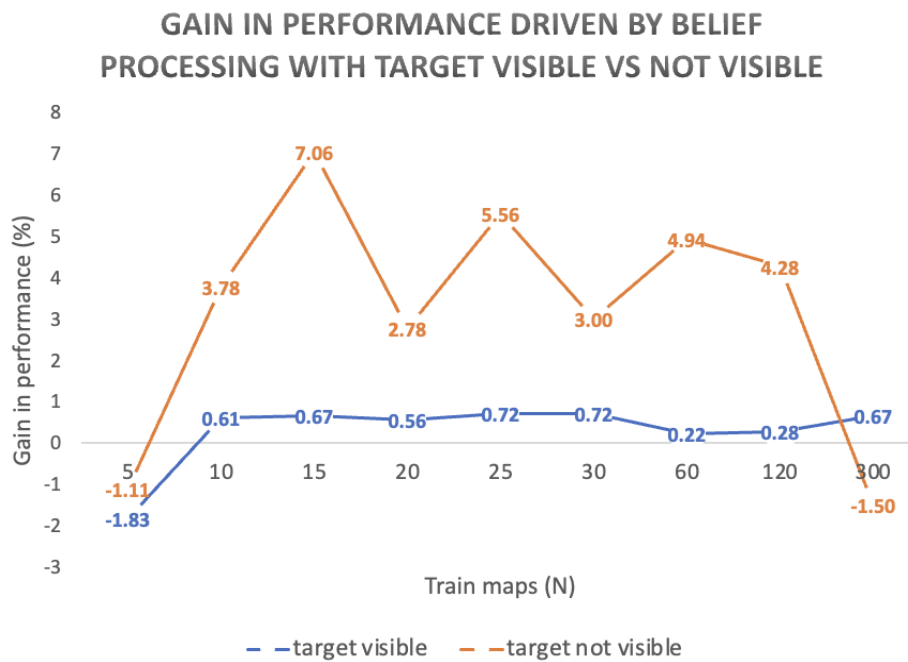

**Figure 5.** Gain in performance driven by the Beliefs head, by number of training maps, in the experimental condition with 3 distractor objects visible and target visible vs not visible to the observed actor.

3.3 Extracting information from avoided distractors

We reasoned that the most informative distractors for predicting an actors' intentional behaviour would be those ignored by the actor during its previous steps. *Figure 6* shows the corresponding performance by both the Beliefs and NoBeliefs architectures in settings with a varying number of ignored objects by the actor (examples provided for 25 and 120 maps used for training). In addition, target visibility (whether the target was in or out of the actor's field of view) was also changed. Independently of the considered architecture or number of training maps, performance was much higher (up to 99.00%) when 3 ignored objects were present in the environment (*Figure 6, A and C*). The worst performance was still high at 79.67% when only 1 object was ignored and the target was not visible to the actor (*Figure 6, D*). The maximum gain of the Beliefs over NoBeliefs architecture was 6.56% (p value < 0.001) when 1 object was ignored and a high number of training maps (120 vs 25 maps) was used (*see Figure 7*). We assume that this reflects a strong contribution of belief learning in discarding objects that are not yet ignored, when these are not being approached in an efficient way by the actor (i.e., the actor is not moving in a goal-directed behaviour as the actor is still searching for the target). We assessed this specific condition in the next experiment. In addition, our results suggest that learning to predict beliefs is useful even when 3 objects have been ignored, especially with few training maps. This implies that the gradient for belief prediction helps learning to discard the ignored objects.

From a ToM perspective, these results are coherent with the developmental trend observed in previous experiments with respect to the impact of beliefs on target prediction. Indeed, when considering the condition in which beliefs were most impactful (i.e., object ignored and target not visible), the gain in performance driven by beliefs processing was highest at a higher level of experience (120 vs 25 maps) (*see Figure 7*). Overall, this suggests that the developmental trend extends to visual crowding when the target is not visible by the actor, as well as when the target is visible. A developmental study by Richardson et al., (2023) investigating ToM in sighted and congenitally blind children (4-17 year old) indeed concluded that vision facilitates, but is not necessary for, ToM development. As an extension, our results support engagement in ToM and its development in scenarios where visual information is not available. Recognizing that the impact of the different environmental conditions is largely significant, it remains challenging for an observer to accurately assess the probability of these environmental conditions. Hence, consistent improvements across all scenarios serve as compelling evidence underscoring the importance of predicting others' beliefs.

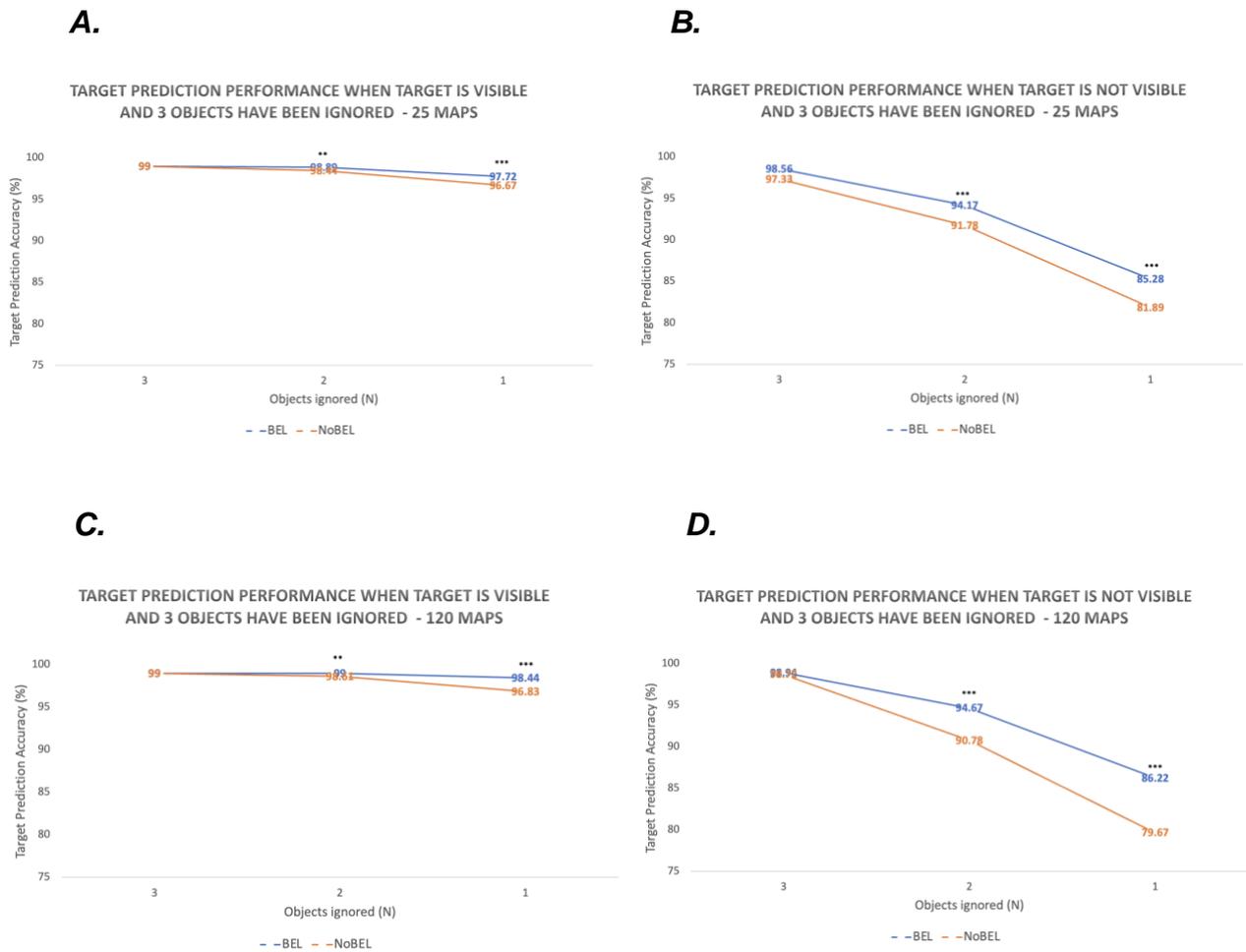

**Figure 6.** Target prediction accuracies for the Beliefs vs NoBeliefs architectures in conditions with varying number of ignored objects and target visible to the actor (25 maps (A), 120 maps (C)) or target not visible to the actor (25 maps (B), 120 maps (D)). Average target prediction accuracy at each number of train maps; *: significant at the .05 level (2-tailed); **: significant at the .01 level (2-tailed); ***: significant at the .001 level (2-tailed).

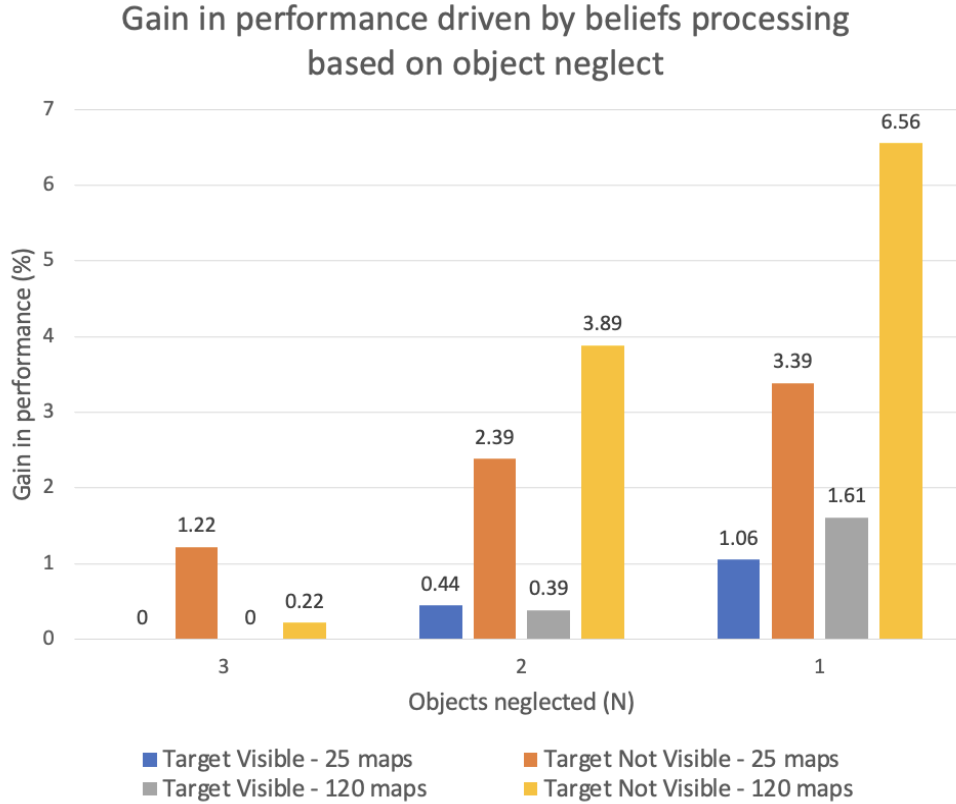

**Figure 7.** Gain in performance driven by beliefs processing based on the number of objects ignored, by target visibility and number of maps.

### 3.4 Extracting information from target-aligned distractors

We assumed that the least informative distractors would be those aligned with the future trajectory of the actor when moving towards the target. Target visibility (whether the target was in or out of the actor's field of view) was also changed. As *Figure 8* shows, the presence of such distractors strongly impacted the system performance (examples provided for 25 and 120 maps used for training). The worst performance (11.56%) was reached by the NoBeliefs architecture when 3 objects were aligned with the target (25 training maps, *Figure 8, B*), thus the target was visible to the actor. In contrast, the Beliefs architecture showed an even stronger gain in this challenging condition, with a top gain of 12.83% over the NoBeliefs architecture (56.67 vs 43.83%, respectively, p value < 0.001) (120 training maps, *Figure 9*), thus a relative gain of about 30% during target-visible conditions. We interpreted this result as evidence that beliefs learning provides a more efficient way of learning to discard distractors that have become visible to the actor before the target itself, even if they are aligned. The Beliefs architecture is likely to better exploit the agent's suboptimal approach behaviour toward visible distractors, even if they are finally traversed. When the target is not visible by the actor, the performance gain of the Belief architecture also reaches a top gain of 11.22% over the NoBelief architecture (46.28 vs 35.06%, respectively, p value < 0.001) (120 training maps, *Figure 9*). This shows that learning to predict beliefs helps learning to interpret beliefs-driven searching behaviours. This is important information for a predicting agent, considering that beliefs-driven searching behaviours would be suboptimal in terms of object approaching, and thus challenging to predict with a teleological approach.

From a ToM perspective, results from sections C and D advanced the previous findings by highlighting that, rather than relying on randomness, the multi-task-induced regularisation between beliefs and target processing is mostly beneficial for recognising others' beliefs-driven behaviours. This in turn allows better disambiguation of objects in the environment, ultimately resulting in improved prediction of others' intentions. While supporting early ToM emergence, this study also highlights the advantages of having a ToM (intended here as beliefs processing) for improved human behaviour prediction in several, and challenging, situations (e.g., when an actor approaches multiple aligned objects). This is in concordance with previous developmental literature suggesting infants' ability to infer an observed agent's preference for one of two objects when the actor approaches them, i.e., condition where target and a distractor object are aligned. For example, Hamlin et al. (2013) investigated 10-month-old infants' judgements in a social evaluation task and indicated infants' ToM ability to attribute to an observed agent preference for an object, as supported by their computational modelling.

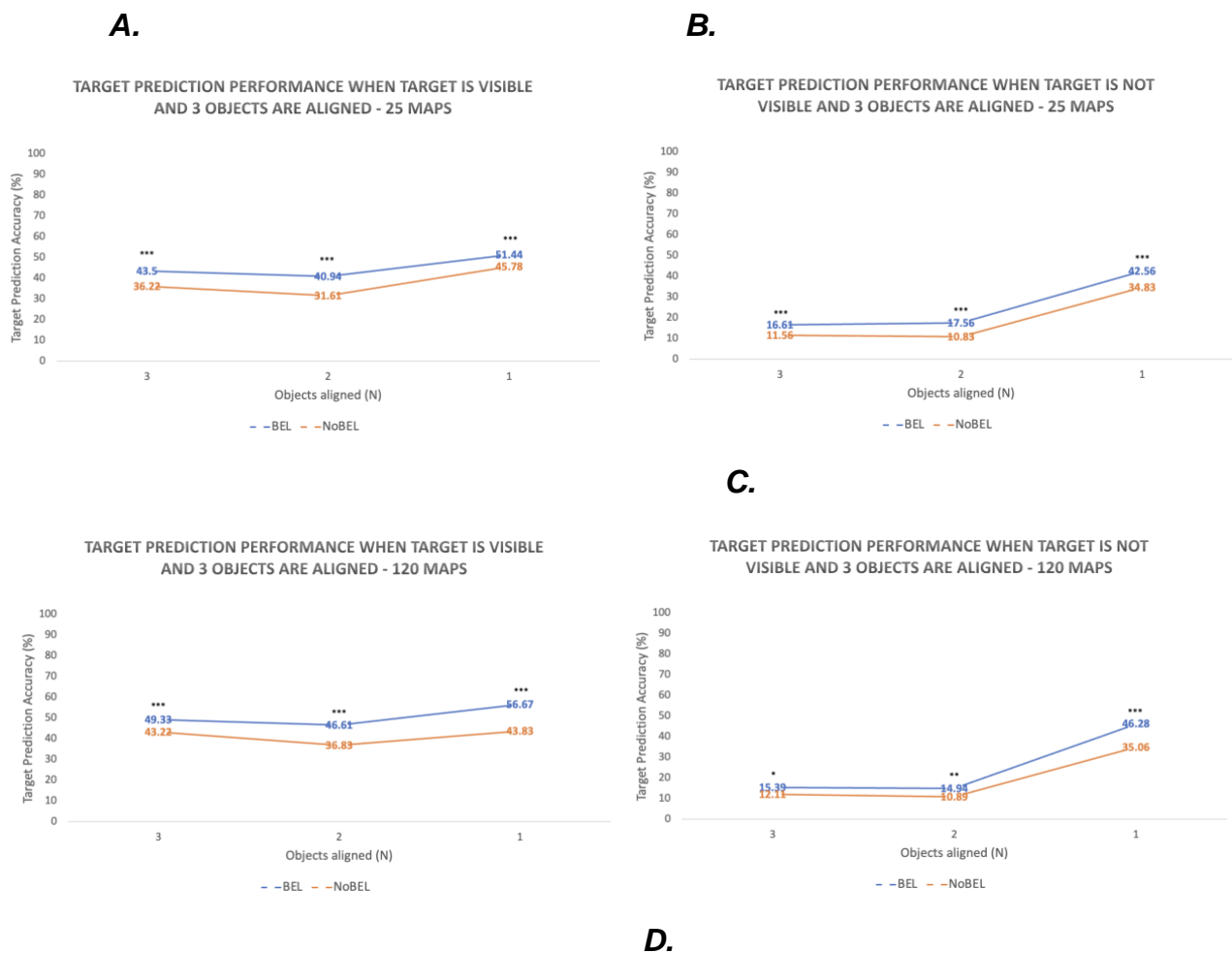

**Figure 8.** Target prediction accuracies for the Beliefs vs NoBeliefs architectures in conditions with varying number of aligned objects to the target, and target visible to the actor (25 maps (A), 120 maps (C)) or target not visible to the actor (25 maps (B), 120 maps (D)). Average target prediction accuracy at each number of train maps; *: significant at the .05 level (2-tailed); **: significant at the .01 level (2-tailed); ***: significant at the .001 level (2-tailed).

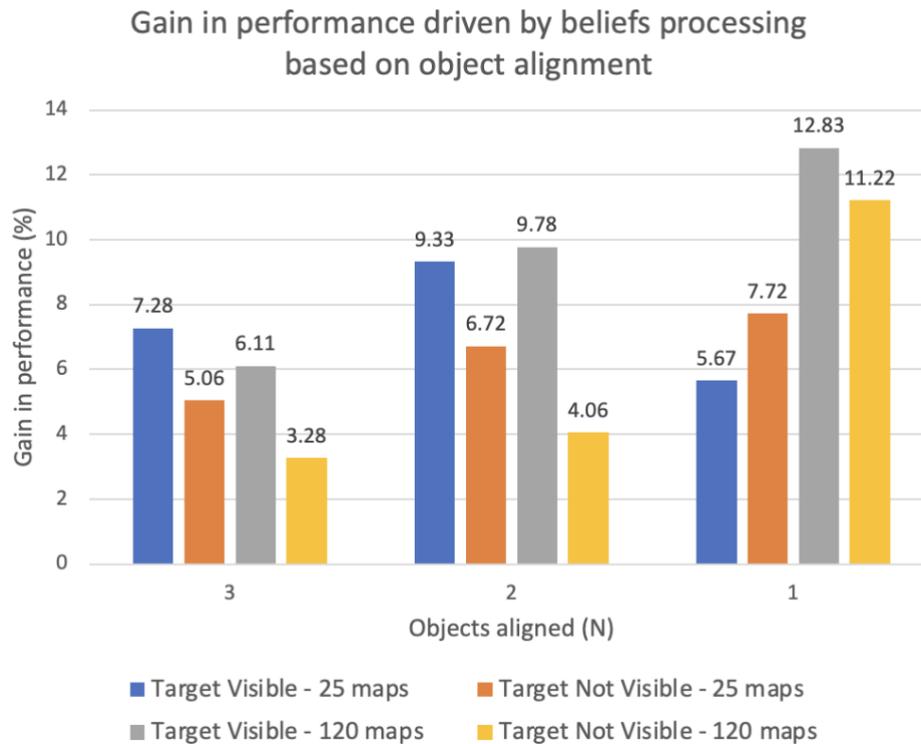

**Figure 9.** Gain in performance driven by beliefs processing based on the number of objects aligned, by target visibility and number of maps.

3.5 Generalisation to different actors

Lastly, we explored whether the advantageous effect of learning to explicitly represent beliefs under the self-observation hypothesis would help interpreting also the behaviour of observed actors with a decision system different to that of the observer. Specifically, whether this synergy between beliefs and intention learning would be useful to interpret behaviours with different modes and effectiveness compared to those used for training. This was tested by changing actors' (1) beliefs representation precision (which could be related to an actor with different cognitive abilities), and (2) action speed (which could be related to an actor with different physical abilities).

Differing beliefs representation precision (1) was achieved by varying the number of samples used by the POMCP algorithm (Silver & Veness, 2010) driving the observed actor's behaviour. More specifically, actors provided at testing differed in the amount of available cognitive resources determining the depth of navigation in their internal model of action-state relationships and associated rewards. Therefore, limited cognitive resources result in incorrect representation and action-sequence assessments, leading to suboptimal deliberation, planning and choices (Ognibene et al., 2019a). While actors provided during training presented high cognitive capabilities (250 max samples), those at testing had lower cognitive capabilities, i.e., 150, 50, 25 and 5 max samples. *Figure 10* shows good target prediction accuracies for actors with varying cognitive capabilities comparable to our original results, regardless of the architecture; although accuracies were lower for actors who were increasingly different from the observer (e.g., 5 max samples condition). Similarly, the role of beliefs for target prediction remains evident, although varying significance was observed. Specifically, a lower gain in performance driven by beliefs processing can be globally reported when observing actors with lower cognitive capabilities.

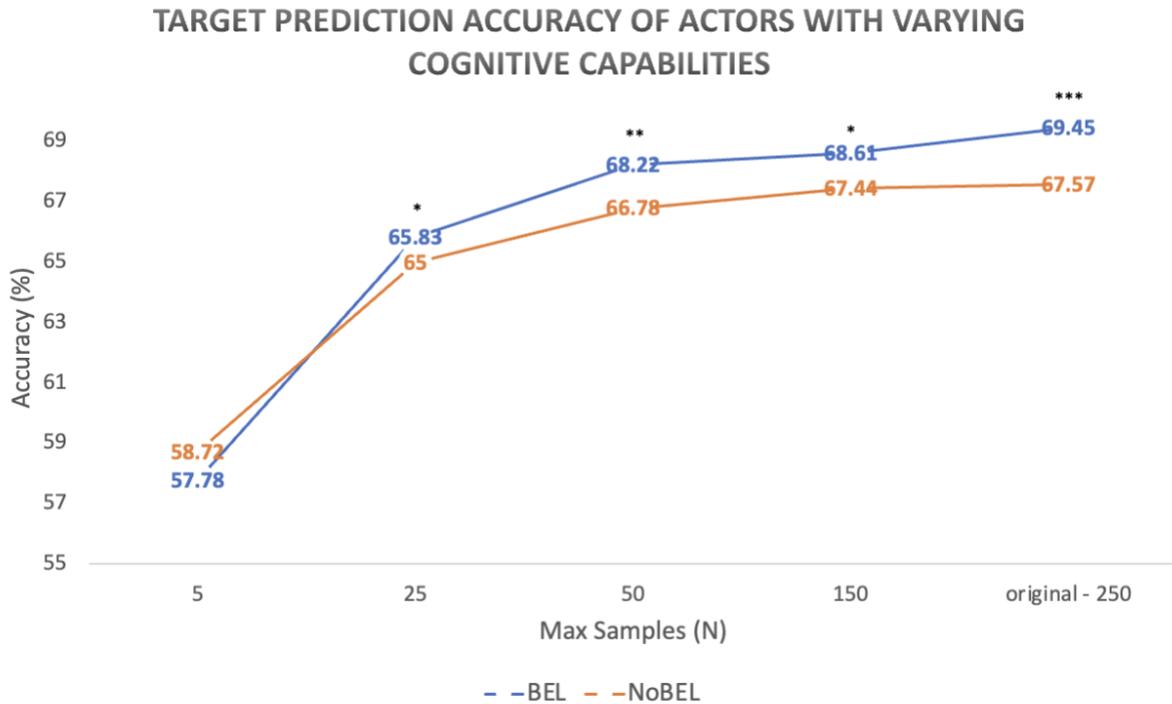

**Figure 10.** Target prediction accuracy for the Beliefs and NoBeliefs architectures based on actors' cognitive capabilities (number of max samples) for representative 25 maps. Average target prediction accuracies are reported. *: significant at the .05 level (2-tailed); **: significant at the .01 level (2-tailed); ***: significant at the .001 level (2-tailed).

Differing physical ability (2) was achieved by testing both the originally trained architectures with simulations which saw actors moving around the grid world at different speeds throughout the task (i.e., at x0.75, x0.9, x1.1, and x1.25 the speed on which the nets were trained upon). *Figure 11* shows that averaged accuracies at testing across maps generally revealed decreased accuracy when predicting slower actors' behaviour, while accuracies remained comparable to the original study when predicting faster actors' behaviours. These results may be driven by the fact that when observing slower actors, especially in the x0.75 condition, the actors are almost still for multiple steps, thus leading to less information and ultimately to lower performances. The opposite is valid with faster actors. Nevertheless, in a post-hoc analysis, we observed that short exposure (i.e., a brief re-training over 5 maps) to a few behaviours of slower actors (x0.75 and x0.9 speed) resulted in improved significance of the Beliefs over NoBeliefs architecture predictive performance (detailed results are omitted from this paper due to space constraints; however, they are available upon request). Therefore, an advantage of beliefs processing towards target prediction was seen in most conditions (statistically significant). We must note that the prediction performance difference is measured over generic behaviours, as in section A, which shows similar performance. Therefore, we can expect a stronger gain of the Beliefs architecture during the observation of beliefs-driven behaviours, as in sections B, C and D.

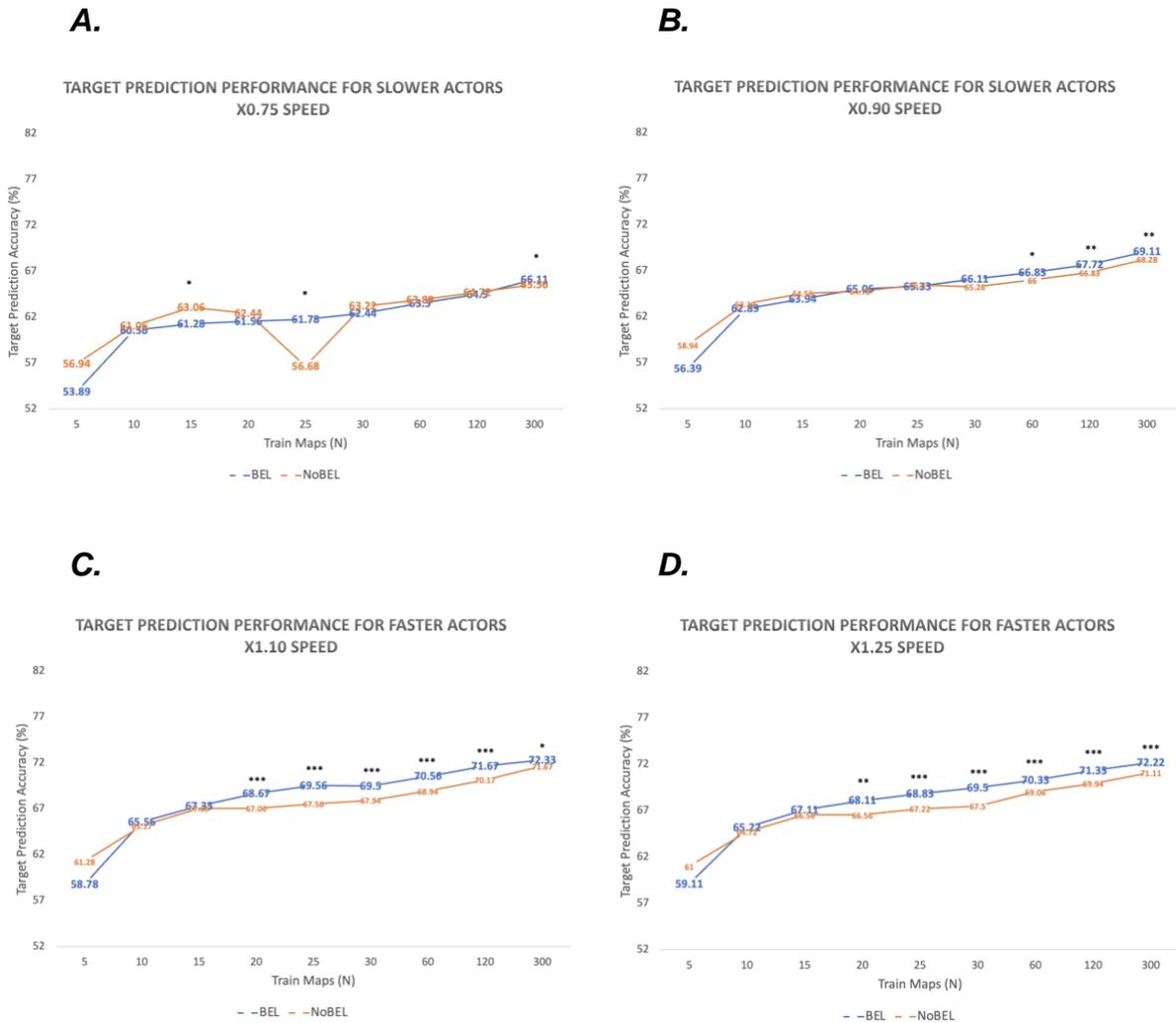

**Figure 11.** Target prediction accuracies for the Beliefs vs NoBeliefs architectures in the conditions with varying complexity of the actor's physical abilities for representative 25 maps. Speed underlying actor's behaviour: (A) x0.75; (B) x0.90; (C) x1.1; (D) x1.25. Average target prediction accuracies are reported. *: significant at the .05 level (2-tailed); **: significant at the .01 level (2-tailed); ***: significant at the .001 level (2-tailed).

Overall, these results suggest an ability of these architectures to generalise to observed actors whose behaviours, driven by their cognitive or physical capabilities, vary and differ from the self. Although lower generalisation is seen in conditions that are increasingly different. Crucially, from a ToM perspective, these results indicate learning to explicitly represent beliefs through the "like them" approach as a resourceful tool for better predicting behaviours of others under complex circumstances. This is in concordance with previous psychology literature suggesting that infants are able to extend beliefs attribution and behaviours prediction to e.g., adults (e.g., Southgate et al., 2009) whose reaching actions may be faster than the infant's abilities, or agents which complete biomechanically impossible actions (thus with a different physical ability to the infant) (e.g., Southgate et al., 2008).

## 4. Conclusions

Studying ToM while taking advantage of the crosstalk between disciplines allows a more global and complete understanding of this human cognition (see also Cangelosi & Schlesinger, 2018; Hassabis et al., 2017; Lake et al., 2017; Sandini et al., 2021). By capitalising on psychology and computational modelling techniques, we provide new insights into human and machine ToM.

In a series of studies, we investigated the "like them" approach to learning explicit beliefs representation for predicting others' intentions and its developmental trajectory. This was done in an attempt to show the advantages and computational sustainability of beliefs processing throughout development towards understanding others' mental states, to inform both human and machine ToM development. Overall, our results show the presence and importance of a developmental synergy of learning beliefs and intention attribution simultaneously, especially when limited samples are available and acquiring new ones is expensive or risky. We show that this architectural solution is particularly helpful in complex but ecologically common situations, such as when multiple distracting objects are present and close in the environment or when interpreting suboptimal exploratory behaviours performed with limited sensory input. We also demonstrate that the acquired ToM model, trained upon self-observation, is able to generalise (in terms of intentions predictions) to different actors, while preserving the intention-beliefs advantageous synergy. Altogether, these results outlined that taking a "like them" approach for beliefs is beneficial towards improved performance for predicting others' intentions. In line with previous computational studies in the field (Baker et a;., 2017; Rabinowitz et al., 2018; Ramırez & Geffner, 2011), we adopted a simplified environment model which can have an impact on models' performance (Ognibene et al., 2019a). Learning to explicitly represent beliefs does indeed increase, although only linearly, the computational demands and complexity of the task. However, it ultimately helps predict others' intentions. This is the case even in a simple feedforward architecture where the impact of beliefs prediction on intention prediction is not hardwired through some specific mechanism but indirectly occurs through the influence of multi-task learning supported by self-observation-based training and its regularisation effect. This is valid when considering varying environments, observers, and actors. We hypothesise that ad hoc architectures can achieve even stronger improvements maybe with lower computational costs.

These results can be interpreted both from a psychology and robotics perspective. From a psychology perspective, our results provide a computational foundation for previous research on ToM emergence and development. Specifically, our results contribute to the literature supporting early ToM emergence (e.g., Southgate and Vernetti, 2014; Hamlin et al., 2013; see also Supplemental Table 1) and indicating a developmental trend for ToM development, i.e., its emergence in infancy (e.g., Southgate and Vernetti, 2014; Hamlin et al., 2013), its further development during adolescence (e.g., Meinhardt-Injac et al., 2020) and middle adulthood (e.g., Spenser et al., 2020) with increasing experience, and finally plateauing or decreasing in late adulthood (e.g., Pearlman-Avnion et al., 2018). Furthermore, our results contribute to the debate surrounding the mechanisms underlying ToM (Baker et al., 2017; Gergely & Csibra, 1997; Hamlin et al., 2013; Keysers & Gazzola, 2007) and provide evidence for the viability of the "like-them" and association mechanisms as likely candidate mechanisms underlying this cognitive ability. As an extension, they also support the notion that individuals need to share a "representational code" for understanding each other (the "like them" assumption) (Decety & Chaminade, 2003), as well as previous developmental literature suggesting that infants may have an altercentric view since birth and only later in development develop a first-person representational perspective (Happé, 2003; Kampis & Kovács, 2022; Southgate, 2020). Importantly, our results support previous literature that indicates that infant successfully complete false-belief tasks because of their ToM capability (e.g., Southgate and Vernetti, 2014; Hamlin et al., 2013; see also Supplemental Table 1), rather than reliance on non-ToM computations (e.g., Heyes, 2014a, 2014b; Perner & Ruffman, 2005); also providing a computational foundation to infants' success at these tasks, which could supplement and revisit failed replication studies of infant ToM abilities for completing false-belief tasks using behavioural measures (Kampis et al., 2021). By showing an increasingly positive impact of belief processing from an early age under circumstances of others' incomplete or false knowledge about the environment, our results point to the presence of the ability to engage in belief attribution for predicting individuals' behaviours in false-belief tasks from infancy. This finding also contributes to the debate surrounding implicit vs explicit ToM (Baillargeon et al., 2016; Vierkant, 2012) by showing that intuitive processes may support the acquisition of belief representation and their use for predicting others' behaviours. Finally, our results indicate that self-other similarity is a factor not as important as previously thought for ToM ability (Rabinowitz et al., 2018), in contrast to e.g., the simulation account previously suggested as a precursor of ToM (e.g., Keysers & Gazzola, 2007). Again, this points to a shared "representational code" (Decety & Chaminade, 2003) which may associate all individuals, while more stringent similarity between self and other may not be a requirement of ToM. In addition, this architectural framework suffers less from interference between the action execution and action perception processes (i.e., simulation theory, Rizzolatti et al., 2001), showing a stronger fit for social context while predicting stronger interferences during learning.

From a robotics perspective, we contributed in terms of computational modelling of ToM and developed a viable architecture for social robots. Specifically, compared to other computational models that focus on predicting beliefs-driven actions (Chen et al., 2021; Nguyen & Gonzalez, 2022; Rabinowitz et al., 2018), we show that learning explicit representation of others' beliefs can improve behaviour prediction performance. More specifically, self-observing own mental states to learn to explicitly represent others' beliefs seems to not introduce limiting biases on behaviour prediction performance (Rabinowitz et al. 2018) and is less computationally complex than previously believed (Apperly & Butterfill, 2009; Baker et al., 2017; Butterfill & Apperly, 2013; Devin & Alami, 2016; Görür et al., 2017). Therefore, learning these skills together seems to introduce a synergy that will resolve in part the limitations associated with the high variability of behaviours in different contexts without increasing the amount of training samples (which are difficult to obtain and ecologically expensive) and bridge the gap between human and AI learning speeds. In terms of a viable architecture for social robotics, this paper proposes a human-inspired robotic architecture which speeds-up the acquisition of socio-cognitive prediction skills, using the "like them" approach extended from the "like me" psychological theory, which strongly improves the interpretation of beliefs-driven behaviours. Furthermore, the proposed architecture is able to generalise beliefs-based behaviour prediction to actors with different cognitive and physical abilities, a fundamental characteristic of our work considering previous research aimed at reducing generalisation errors in deep learning (e.g., Zhang et al., 2021). The contributions of these experimental studies can inform the development of artificial architectures and social agents equipped with a machine ToM, contributing to one of the "Grand Challenges of Science Robotics" (Yang et al., 2018, p. 9). While our experiments were conducted in a relatively simple gridworld environment (11x11 gridworld), the extensive evaluation highlights unexpected and counterintuitive results that warrant further validation in more complex and realistic settings to increase transferability to real-life scenarios. Additional experiments could test whether this approach could be extended further to improve the meta-learning method to extract richer agent-specific representations (embeddings) online as proposed in (Rabinowitz et al. 2018).

In conclusion, our computational approach reveals a developmental synergy between intentions and beliefs representations. Through our experiments, we demonstrate the advantageous impact of explicitly representing beliefs on behaviour prediction accuracy, in several scenarios and when generalising to diverse actors. This paper contributes to advancing both theoretical understanding and practical applications of ToM in human cognition and machine learning, with implications for addressing key challenges in social robotics.

## 5. Conflicts of interest

The authors have no conflicts of interest to disclose.

## 6. Author contributions

*Dr Francesca Bianco:* Conceptualization, Data curation, Formal analysis, Investigation, Methodology, Software, Visualization, Writing - original draft and final manuscript; *Dr Dimitri Ognibene:* Conceptualization, Methodology, Software, Supervision and Writing - review & editing; *Dr Silvia Rigato:* Supervision and Writing - review & editing; *Dr Maria Laura Filippetti:* Supervision and Writing - review & editing.

## 7. Acknowledgments

Dr Francesca Bianco and Dr Dimitri Ognibene have been supported by European Union's Horizon 2020 research and innovation programme under grant agreement number 824153 "POTION".

# References

Abubshait, A., & Wiese, E. (2017). You Look Human, But Act Like a Machine: Agent Appearance and Behavior Modulate Different Aspects of Human–Robot Interaction. *Frontiers in Psychology, 8,* 1393. https://doi.org/10.3389/fpsyg.2017.01393

Anderson, J. R., & Lebiere, C. J. (2014). *The atomic components of thought.* New York, NY: Psychology Press. https://doi.org/10.4324/9781315805696

Apperly, I. A., & Butterfill, S. A. (2009). Do humans have two systems to track beliefs and belief-like states? *Psychological Review, 116(4)*, 953–970. https://doi.org/10.1037/a0016923

Asakura, N., & Inui, T. (2016). A Bayesian Framework for False Belief Reasoning in Children: A Rational Integration of Theory-Theory and Simulation Theory. *Frontiers in Psychology, 7*. https://www.frontiersin.org/article/10.3389/fpsyg.2016.02019

Baillargeon, R., Scott, R. M., & Bian, L. (2016). Psychological reasoning in infancy. *Annual review of psychology, 67*, 159-186. https://doi.org/10.1146/annurev-psych-010213-115033

Baird, J. A., & Astington, J. W. (2005). The development of the intention concept: From the observable world to the unobservable mind. *The new unconscious*, 256-276.

Baron-Cohen, S., Leslie, A. M., & Frith, U. (1985). Does the autistic child have a "theory of mind"? *Cognition, 21(1),* 37–46. https://doi.org/10.1016/0010-0277(85)90022-8

Bhat, A. A., Mohan, V., Sandini, G., & Morasso, P. (2016). Humanoid infers Archimedes' principle: Understanding physical relations and object affordances through cumulative learning experiences. *Journal of The Royal Society Interface, 13(120),* 20160310. https://doi.org/10.1098/rsif.2016.0310

Bianco, F., & Ognibene, D. (2019). Transferring adaptive theory of mind to social robots: Insights from developmental psychology to robotics. *In Social Robotics: 11th International Conference, ICSR 2019, Madrid, Spain, November 26–29, 2019, Proceedings 11* (pp. 77-87). Springer International Publishing. https://doi.org/10.1007/978-3-030-35888-4_8

Bianco, F., & Ognibene, D. (2020). From Psychological Intention Recognition Theories to Adaptive Theory of Mind for Robots: Computational Models. *Companion of the 2020 ACM/IEEE International Conference on Human-Robot Interaction,* 136–138. https://doi.org/10.1145/3371382.3378364

Baker, C. L., Jara-Ettinger, J., Saxe, R., & Tenenbaum, J. B. (2017). Rational quantitative attribution of beliefs, desires and percepts in human mentalizing. *Nature Human Behaviour, 1(4),* 1–10. https://doi.org/10.1038/s41562-017-0064

Buccino, G., Lui, F., Canessa, N., Patteri, I., Lagravinese, G., Benuzzi, F., Porro, C. A., & Rizzolatti, G. (2004). Neural circuits involved in the recognition of actions performed by nonconspecifics: An FMRI study. *Journal of Cognitive Neuroscience, 16(1),* 114–126. https://doi.org/10.1162/089892904322755601

Butterfill, S. A., & Apperly, I. A. (2013). How to Construct a Minimal Theory of Mind. *Mind & Language, 28(5),* 606–637. https://doi.org/10.1111/mila.12036

Cangelosi, A., & Schlesinger, M. (2018). From Babies to Robots: The Contribution of Developmental Robotics to Developmental Psychology. *Child Development Perspectives, 12(3),* 183–188. https://doi.org/10.1111/cdep.12282

Chen, B., Vondrick, C., & Lipson, H. (2021). Visual behavior modelling for robotic theory of mind. *Scientific Reports, 11(1),* 424. https://doi.org/10.1038/s41598-020-77918-x

Crawshaw, M. (2020). Multi-Task Learning with Deep Neural Networks: A Survey. *arXiv preprint arXiv:2009.09796* [Cs, Stat]. http://arxiv.org/abs/2009.09796

Csibra, G., & Gergely, G. (2007). 'Obsessed with goals': Functions and mechanisms of teleological interpretation of actions in humans. *Acta Psychologica, 124(1),* 60–78. https://doi.org/10.1016/j.actpsy.2006.09.007

Decety, J., & Chaminade, T. (2003). When the self represents the other: A new cognitive neuroscience view on psychological identification. *Consciousness and Cognition, 12(4),* 577–596. https://doi.org/10.1016/S1053-8100(03)00076-X

Demiris, Y. (2007). Prediction of intent in robotics and multi-agent systems. *Cognitive Processing, 8(3),* 151–158. https://doi.org/10.1007/s10339-007-0168-9

Demiris*, Y., & Johnson, M. (2003). Distributed, predictive perception of actions: a biologically inspired robotics architecture for imitation and learning. *Connection science, 15(4),* 231-243. https://doi.org/10.1080/09540090310001655129

Demiris, Y., & Khadhouri, B. (2006). Hierarchical attentive multiple models for execution and recognition of actions. *Robotics and Autonomous Systems, 54(5),* 361–369. https://doi.org/10.1016/j.robot.2006.02.003

Devaine, M., Hollard, G., & Daunizeau, J. (2014). The social Bayesian brain: Does mentalizing make a difference when we learn? *PLoS Computational Biology, 10(12)*, e1003992.

Devin, S., & Alami, R. (2016). *An Implemented Theory of Mind to Improve Human-Robot Shared Plans Execution.* The Eleventh ACM/IEEE International Conference on Human Robot Interaction, 319–326. https://doi.org/10.1109/HRI.2016.7451768

De Bruin, L. C., & Newen, A. (2012). An association account of false belief understanding. *Cognition, 123(2),* 240–259. https://doi.org/10.1016/j.cognition.2011.12.016

Friston, K., Rigoli, F., Ognibene, D., Mathys, C., Fitzgerald, T., & Pezzulo, G. (2015). Active inference and epistemic value. *Cognitive neuroscience, 6(4),* 187-214. https://doi.org/10.1080/17588928.2015.1020053

Frith, C. D., & Frith, U. (1999). Interacting Minds—A Biological Basis. Science, 286(5445), 1692–1695. https://doi.org/10.1126/science.286.5445.1692

Frith, C. D.,, & Frith, U. (2005). Theory of mind. *Current Biology, 15(17),* R644–R645. 10.1016/j.cub.2005.08.041

Frith, C. D., & Frith, U. (2006). The neural basis of mentalizing. *Neuron, 50(4),* 531–534. https://doi.org/10.1016/j.neuron.2006.05.001

Gergely, G., & Csibra, G. (1997). Teleological reasoning in infancy: The infant's naive theory of rational action. A reply to Premack and Premack. *Cognition, 63(2),* 227–233. https://doi.org/10.1016/s0010-0277(97)00004-8

Giese, M. A., & Rizzolatti, G. (2015). Neural and Computational Mechanisms of Action Processing: Interaction between Visual and Motor Representations. *Neuron, 88(1),* 167–180. https://doi.org/10.1016/j.neuron.2015.09.040

Gonzalez, C., Lerch, J. F., & Lebiere, C. (2003). Instance-based learning in dynamic decision making. *Cognitive Science, 27(4),* 591–635. https://doi.org/10.1207/s15516709cog2704_2

Görür, O., Rosman, B., Hoffman, G., & Albayrak, S. (2017, March 6). *Toward Integrating Theory of Mind into Adaptive Decision-Making of Social Robots to Understand Human Intention.* http://hdl.handle.net/10204/9653

Hamlin, J. K., Ullman, T., Tenenbaum, J., Goodman, N., & Baker, C. (2013). The mentalistic basis of core social cognition: Experiments in preverbal infants and a computational model. *Developmental Science, 16(2),* 209–226. https://doi.org/10.1111/desc.12017

Happé, F. (2003). Theory of mind and the self. *Annals of the New York Academy of Sciences, 1001(1),* 134–144.

Hassabis, D., Kumaran, D., Summerfield, C., & Botvinick, M. (2017). Neuroscience-Inspired Artificial Intelligence. *Neuron, 95(2),* 245–258. https://doi.org/10.1016/j.neuron.2017.06.011

He, H., Boyd-Graber, J., Kwok, K., & Daumé III, H. (2016). Opponent Modeling in Deep Reinforcement Learning. *arXiv preprint arXiv:1609.05559* [Cs]. http://arxiv.org/abs/1609.05559

Hegel, F., Krach, S., Kircher, T., Wrede, B., & Sagerer, G. (2008). *Theory of mind (ToM) on robots: A functional neuroimaging study.* In Proceedings of the 3rd ACM/IEEE international conference on Human robot interaction (pp. 335-342).

Heyes, C. (2010). Where do mirror neurons come from? *Neuroscience and Biobehavioral Reviews, 34(4)*, 575–583. https://doi.org/10.1016/j.neubiorev.2009.11.007

Heyes, C. (2014a). Submentalizing: I Am Not Really Reading Your Mind. *Perspectives on Psychological Science: A Journal of the Association for Psychological Science, 9(2),* 131–143. https://doi.org/10.1177/1745691613518076

Heyes, C. (2014b). False belief in infancy: A fresh look. *Developmental Science, 17(5),* 647–659. https://doi.org/10.1111/desc.12148

Hiatt, L., Harrison, A., & Trafton, J. (2011). *Accommodating Human Variability in Human-Robot Teams through Theory of Mind.* In Twenty-second international joint conference on artificial intelligence. https://doi.org/10.5591/978-1-57735-516-8/IJCAI11-345

Hoffmann, M., Chinn, L. K., Somogyi, E., Heed, T., Fagard, J., Lockman, J. J., & O'Regan, J. K. (2017). *Development of reaching to the body in early infancy: From experiments to robotic models.* 2017 Joint IEEE International Conference on Development and Learning and Epigenetic Robotics (ICDL-EpiRob), 112–119. https://doi.org/10.1109/DEVLRN.2017.8329795

Hommel, B., Müsseler, J., Aschersleben, G., & Prinz, W. (2001). The Theory of Event Coding (TEC): A framework for perception and action planning. *Behavioral and Brain Sciences, 24(5),* 849–878. https://doi.org/10.1017/S0140525X01000103

Hyde, D., Simon, C., Ting, F., & Nikolaeva, J. (2018). Functional Organization of the Temporal-Parietal Junction for Theory of Mind in Preverbal Infants: A Near-Infrared Spectroscopy Study. *The Journal of Neuroscience, 38,* 0264–17. https://doi.org/10.1523/JNEUROSCI.0264-17.2018

Jaderberg, M., Mnih, V., Czarnecki, W. M., Schaul, T., Leibo, J. Z., Silver, D., & Kavukcuoglu, K. (2017). *Reinforcement learning with unsupervised auxiliary tasks.* In International Conference on Learning Representations.

James, W. (1890). *The principles of psychology*, Vol I. (pp. xii, 697). Henry Holt and Co. https://doi.org/10.1037/10538-000

Kampis, D., Karman, P., Csibra, G., Southgate, V., & Hernik, M. (2021). A two-lab direct replication attempt of Southgate, Senju and Csibra (2007). *Royal Society open science, 8(8),* 210190. https://doi.org/10.1098/rsos.210190

Kampis, D., & Kovács, Á. M. (2022). Seeing the world from others' perspective: 14-month-olds show altercentric modulation effects by others' beliefs. *Open Mind, 1–19.* https://doi.org/10.1162/opmi_a_00050

Kampis, D., Parise, E., Csibra, G., & Kovács, Á. M. (2015). Neural signatures for sustaining object representations attributed to others in preverbal human infants. *Proceedings of the Royal Society B: Biological Sciences, 282(1819),* 20151683. https://doi.org/10.1098/rspb.2015.1683

Keysers, C., & Gazzola, V. (2007). Integrating simulation and theory of mind: From self to social cognition. *Trends in Cognitive Sciences, 11(5),* 194–196. https://doi.org/10.1016/j.tics.2007.02.002

Knox, W. B., Otto, A. R., Stone, P., & Love, B. C. (2012). The nature of belief-directed exploratory choice in human decision-making. *Frontiers in psychology, 2,* 398. https://doi.org/10.3389/fpsyg.2011.00398

Koay, K. L., Sisbot, E. A., Syrdal, D. S., Walters, M. L., Dautenhahn, K., & Alami, R. (2007). *Exploratory study of a robot approaching a person in the context of handing over an object.* In AAAI spring symposium: multidisciplinary collaboration for socially assistive robotics (pp. 18-24).

Kominis, F., & Geffner, H. (2015). Beliefs in Multiagent Planning: From One Agent to Many. *Proceedings of the International Conference on Automated Planning and Scheduling, 25(1),* 147–155. https://doi.org/10.1609/icaps.v25i1.13726

Kominis, F., & Geffner, H. (2017). Multiagent Online Planning with Nested Beliefs and Dialogue. *Proceedings of the International Conference on Automated Planning and Scheduling, 27,* 186–194. https://doi.org/10.1609/icaps.v27i1.13826

Kosinski, M. (2023). Evaluating Large Language Models in Theory of Mind Tasks. *arXiv e-prints, arXiv-2302.02083.* https://doi.org/10.48550/arXiv.2302.02083

Kovács, Á. M., Téglás, E., & Endress, A. D. (2010). *The social sense: Susceptibility to others' beliefs in human infants and adults.* Science (New York, N.Y.), 330(6012), 1830–1834. https://doi.org/10.1126/science.1190792

Lake, B. M., Ullman, T. D., Tenenbaum, J. B., & Gershman, S. J. (2017). Building machines that learn and think like people. *Behavioral and Brain Sciences,* 40. https://doi.org/10.1017/S0140525X16001837

Langley, C., Cirstea, B. I., Cuzzolin, F., & Sahakian, B. J. (2022). Theory of mind and preference learning at the interface of cognitive science, neuroscience, and AI: A review. *Frontiers in artificial intelligence, 5,* 62. https://doi.org/10.3389/frai.2022.778852

Luo, Y. (2011). Do 10-month-old infants understand others' false beliefs? *Cognition, 121(3),* 289–298. https://doi.org/10.1016/j.cognition.2011.07.011

Ma, X., Gao, L., & Xu, Q. (2023). Tomchallenges: A principle-guided dataset and diverse evaluation tasks for exploring theory of mind. *arXiv preprint arXiv:2305.15068.*

Meinhardt-Injac, B., Daum, M. M., & Meinhardt, G. (2020). Theory of mind development from adolescence to adulthood: Testing the two-component model. *British Journal of Developmental Psychology, 38(2),* 289-303. https://doi.org/10.1111/bjdp.1232029

Meltzoff, A. N. (2007a). 'Like me': A foundation for social cognition. *Developmental Science, 10(1),* 126–134. https://doi.org/10.1111/j.1467-7687.2007.00574.x

Meltzoff, A. N. (2007b). The 'like me' framework for recognizing and becoming an intentional agent. *Acta Psychologica, 124(1),* 26–43. https://doi.org/10.1016/j.actpsy.2006.09.005

Milliez, G., Warnier, M., Clodic, A., & Alami, R. (2014, August). *A framework for endowing an interactive robot with reasoning capabilities about perspective-taking and belief management.* In The 23rd IEEE international symposium on robot and human interactive communication (pp. 1103-1109). IEEE. https://doi.org/10.1109/ROMAN.2014.6926399

Monroy, C., Gerson, S., & Hunnius, S. (2017). Infants' Motor Proficiency and Statistical Learning for Actions. *Frontiers in Psychology, 8.* https://www.frontiersin.org/article/10.3389/fpsyg.2017.02174

Moriguchi, Y., Ban, M., Osanai, H., & Uchiyama, I. (2018). Relationship between implicit false belief understanding and role play: Longitudinal study. *European Journal of Developmental Psychology, 15(2),* 172–183. https://doi.org/10.1080/17405629.2017.1280022

Murray, K., Johnston, K., Cunnane, H., Kerr, C., Spain, D., Gillan, N., Hammond, N., Murphy, D., & Happé, F. (2017). A new test of advanced theory of mind: The 'Strange Stories Film Task' captures social processing differences in adults with autism spectrum disorders. *Autism Research: Official Journal of the International Society for Autism Research, 10(6),* 1120–1132. https://doi.org/10.1002/aur.1744

Nguyen, T. N., & Gonzalez, C. (2022). Theory of mind from observation in cognitive models and humans. *Topics in cognitive science, 14(4),* 665-686. https://doi.org/10.1111/tops.12553

Ognibene, D., & Demiris, Y. (2013). Towards active event recognition. Proceedings of the Twenty-Third International Joint Conference on Artificial Intelligence (pp. 2495-2501).

Ognibene, D., Fiore, V. G., & Gu, X. (2019a). Addiction beyond pharmacological effects: The role of environment complexity and bounded rationality. *Neural Networks : The Official Journal of the International Neural Network Society, 116,* 269–278. https://doi.org/10.1016/j.neunet.2019.04.022

Ognibene, D., Mirante, L., & Marchegiani, L. (2019b). *Proactive intention recognition for joint human-robot search and rescue missions through monte-carlo planning in pomdp environments.* In Social Robotics: 11th International Conference, ICSR 2019, Madrid, Spain, November 26–29, 2019, Proceedings 11 (pp. 332-343). Springer International Publishing.

Onishi, K. H., & Baillargeon, R. (2005). Do 15-Month-Old Infants Understand False Beliefs? *Science (New York, N.y.), 308(5719),* 255–258. https://doi.org/10.1126/science.1107621

Patacchiola, M., & Cangelosi, A. (2016). *A developmental Bayesian model of trust in artificial cognitive systems.* 2016 Joint IEEE International Conference on Development and Learning

and Epigenetic Robotics (ICDL-EpiRob), 117–123. https://doi.org/10.1109/DEVLRN.2016.7846801

Pearlman-Avnion, S., Ron, N., & Ezekiel, S. (2018). Ageing and theory of mind abilities: The benefits of social interaction. *Educational Gerontology, 44(5-6),* 368-377. https://doi.org/10.1080/03601277.2018.1480130

Perner, J., & Ruffman, T. (2005). Infants' Insight into the Mind: How Deep? *Science, 308(5719),* 214–216. https://doi.org/10.1126/science.1111656

Priewasser, B., Rafetseder, E., Gargitter, C., & Perner, J. (2018). Helping as an early indicator of a theory of mind: Mentalism or Teleology? *Cognitive Development, 46,* 69–78. https://doi.org/10.1016/j.cogdev.2017.08.002

Rabinowitz, N., Perbet, F., Song, F., Zhang, C., Eslami, S. M. A., & Botvinick, M. (2018). *Machine Theory of Mind.* Proceedings of the 35th International Conference on Machine Learning, 4218–4227. https://proceedings.mlr.press/v80/rabinowitz18a.html

Raileanu, R., Denton, E., Szlam, A., & Fergus, R. (2018). Modeling Others using Oneself in Multi-Agent Reinforcement Learning. *arXiv preprint arXiv:1802.09640* [Cs]. http://arxiv.org/abs/1802.09640

Rakoczy, H. (2017). *Theory of mind.* In B. Hopkins, E. Geangu, & S. Linkenauger (Eds.), The Cambridge Encyclopedia of Child Development (2nd ed., pp. 505–512). Cambridge University Press. https://doi.org/10.1017/9781316216491.081

Ramırez, M., & Geffner, H. (2011). *Goal recognition over POMDPs: Inferring the intention of a POMDP agent.* In IJCAI (pp. 2009-2014). IJCAI/AAAI. https://doi.org/10.5591/978-1-57735-516-8/IJCAI11-335

Richardson, H., Saxe, R., & Bedny, M. (2023). Neural correlates of theory of mind reasoning in congenitally blind children. *Developmental Cognitive Neuroscience, 63,* 101285. https://doi.org/10.1016/j.dcn.2023.101285

Rizzolatti, G., Fogassi, L., & Gallese, V. (2001). Neurophysiological mechanisms underlying the understanding and imitation of action. *Nature Reviews Neuroscience, 2(9),* 661–670. https://doi.org/10.1038/35090060

Rossi, S., Coppola, A., Gaita, M., & Rossi, A. (2023). *Human–Robot Interaction Video Sequencing Task (HRIVST) for Robot's Behavior Legibility.* IEEE Transactions on Human-Machine Systems. https://ieeexplore.ieee.org/document/10317817

Ruder, S. (2017). An Overview of Multi-Task Learning in Deep Neural Networks. *arXiv preprint arXiv:1706.05098* [Cs, Stat]. http://arxiv.org/abs/1706.05098

Ruffman, T., & Perner, J. (2005). Do infants really understand false belief?: Response to Leslie. *Trends in Cognitive Sciences, 9(10),* 462–463. https://doi.org/10.1016/j.tics.2005.08.001

Sandini, G., Sciutti, A., & Vernon, D. (2021). *Cognitive Robotics*. In M. H. Ang, O. Khatib, & B. Siciliano (Eds.), Encyclopedia of Robotics (pp. 1–7). Springer Berlin Heidelberg. https://doi.org/10.1007/978-3-642-41610-1_198-1

Sap, M., LeBras, R., Fried, D., & Choi, Y. (2022). Neural theory-of-mind? on the limits of social intelligence in large lms. *arXiv preprint arXiv:2210.13312.*

Scarf, D., Imuta, K., Colombo, M., & Hayne, H. (2012). Golden Rule or valence matching? Methodological problems in Hamlin et al. *Proceedings of the National Academy of Sciences, 109(22)*, E1426-E1426.

Schneider, D., Slaughter, V. P., Bayliss, A. P., & Dux, P. E. (2013). A temporally sustained implicit theory of mind deficit in autism spectrum disorders. *Cognition, 129(2),* 410–417. https://doi.org/10.1016/j.cognition.2013.08.004

Scott, R. M., Baillargeon, R., Song, H., & Leslie, A. M. (2010). Attributing false beliefs about non-obvious properties at 18 months. *Cognitive Psychology, 61(4),* 366–395. https://doi.org/10.1016/j.cogpsych.2010.09.001

Scott, M. R., Richman, J. C., & Baillargeon, R. (2015). Infants understand deceptive intentions to implant false beliefs about identity: New evidence for early mentalistic reasoning. *Cognitive Psychology, 82,* 32–56. https://doi.org/10.1016/j.cogpsych.2015.08.003

Senju, A., Southgate, V., Snape, C., Leonard, M., & Csibra, G. (2011). Do 18-months-olds really attribute mental states to others? A critical test. *Psychological Science, 22(7),* 878–880. https://doi.org/10.1177/0956797611411584

Silver, D., & Veness, J. (2010). Monte-Carlo planning in large POMDPs. *Advances in neural information processing systems, 23.*

Sisbot, E. A., Marin-Urias, L. F., Alami, R., & Simeon, T. (2007). A Human Aware Mobile Robot Motion Planner. *IEEE Transactions on Robotics, 23(5),* 874–883. https://doi.org/10.1109/TRO.2007.904911

Skerry, A. E., Carey, S. E., & Spelke, E. S. (2013). First-person action experience reveals sensitivity to action efficiency in prereaching infants. *Proceedings of the National Academy of Sciences of the United States of America, 110(46)*, 18728–18733. https://doi.org/10.1073/pnas.1312322110

Sodian, B., & Kristen, S. (2016). *Theory of Mind.* In Handbook of epistemic cognition (pp. 68–85).

Sodian, B., Kristen-Antonow, S., & Kloo, D. (2020). How Does Children's Theory of Mind Become Explicit? A Review of Longitudinal Findings. *Child Development Perspectives, 14(3)*, 171–177. https://doi.org/10.1111/cdep.12381

Song, H., & Baillargeon, R. (2008). Infants' Reasoning About Others' False Perceptions. *Developmental Psychology, 44(6),* 1789–1795. https://doi.org/10.1037/a0013774

Soubki, A., Murzaku, J., Jordehi, A. Y., Zeng, P., Markowska, M., Mirroshandel, S. A., & Rambow, O. (2024). Views Are My Own, But Also Yours: Benchmarking Theory of Mind using Common Ground. *arXiv preprint arXiv:2403.02451.*

Southgate, V. (2020). Are infants altercentric? The other and the self in early social cognition. *Psychological Review, 127(4),* 505–523. https://doi.org/10.1037/rev0000182

Southgate, V., Johnson, M. H., & Csibra, G. (2008). Infants attribute goals even to biomechanically impossible actions. *Cognition, 107(3),* 1059–1069. https://doi.org/10.1016/j.cognition.2007.10.002

Southgate, V., Johnson, M. H., Osborne, T., & Csibra, G. (2009). Predictive motor activation during action observation in human infants. *Biology Letters, 5(6),* 769–772. https://doi.org/10.1098/rsbl.2009.0474

Southgate, V., Senju, A., & Csibra, G. (2007). Action Anticipation Through Attribution of False Belief by 2-Year-Olds. *Psychological Science, 18(7),* 587–592. https://doi.org/10.1111/j.1467-9280.2007.01944.x

Southgate, V., & Vernetti, A. (2014). Belief-based action prediction in preverbal infants. *Cognition, 130(1),* 1–10. https://doi.org/10.1016/j.cognition.2013.08.008

Spenser, K., Bull, R., Betts, L., & Winder, B. (2020). Underpinning prosociality: Age related performance in theory of mind, empathic understanding, and moral reasoning. *Cognitive Development, 56,* 100928. https://doi.org/10.1016/j.cogdev.2020.100928

Strachan, J., Albergo, D., Borghini, G., Pansardi, O., Scaliti, E., Rufo, A., ... & Becchio, C. (2023). Testing Theory of Mind in GPT Models and Humans. *preprint* https://doi.org/10.21203/rs.3.rs-3262385/v1

Surian, L., Caldi, S., & Sperber, D. (2007). Attribution of beliefs by 13-month-old infants. *Psychological Science, 18(7),* 580–586. https://doi.org/10.1111/j.1467-9280.2007.01943.x

Thellman, S., De Graaf, M., & Ziemke, T. (2022). Mental state attribution to robots: A systematic review of conceptions, methods, and findings. *ACM Transactions on Human-Robot Interaction (THRI), 11(4),* 1-51.

Tomasello, M., Carpenter, M., Call, J., Behne, T., & Moll, H. (2005). Understanding and sharing intentions: The origins of cultural cognition. *Behavioral and Brain Sciences, 28(5),* 675–691. https://doi.org/10.1017/S0140525X05000129

Träuble, B., Marinović, V., & Pauen, S. (2010). Early Theory of Mind Competencies: Do Infants Understand Others' Beliefs? *Infancy: The Official Journal of the International Society on Infant Studies, 15(4),* 434–444. https://doi.org/10.1111/j.1532-7078.2009.00025.x

van Overwalle, F., & Vandekerckhove, M. (2013). Implicit and explicit social mentalizing: Dual processes driven by a shared neural network. *Frontiers in Human Neuroscience, 7*. https://doi.org/10.3389/fnhum.2013.00560

Vierkant, T. (2012). Self-knowledge and knowing other minds: The implicit/explicit distinction as a tool in understanding theory of mind. British Journal of Developmental Psychology, 30(1), 141-155. https://doi.org/10.1111/j.2044-835X.2011.02068.x

Wellman, H. M., Cross, D., & Watson, J. (2001). Meta-analysis of theory-of-mind development: The truth about false belief. *Child Development, 72(3),* 655–684. https://doi.org/10.1111/1467-8624.00304

Wimmer, H., & Perner, J. (1983). Beliefs about beliefs: Representation and constraining function of wrong beliefs in young children's understanding of deception. *Cognition, 13(1),* 103–128. https://doi.org/10.1016/0010-0277(83)90004-5

Winfield, A. F. T. (2018). Experiments in Artificial Theory of Mind: From Safety to Story-Telling. *Frontiers in Robotics and AI, 5*. https://www.frontiersin.org/article/10.3389/frobt.2018.00075

Wolpert, D. M., Doya, K., & Kawato, M. (2003). A unifying computational framework for motor control and social interaction. Philosophical Transactions of the Royal Society of London. *Series B: Biological Sciences, 358(1431),* 593-602. https://doi.org/10.1098/rstb.2002.1238

Woodward, A. L. (1998). Infants selectively encode the goal object of an actor's reach. *Cognition, 69(1),* 1–34. https://doi.org/10.1016/S0010-0277(98)00058-4

Yang, G.-Z., Bellingham, J., Dupont, P. E., Fischer, P., Floridi, L., Full, R., Jacobstein, N., Kumar, V., McNutt, M., Merrifield, R., Nelson, B. J., Scassellati, B., Taddeo, M., Taylor, R., Veloso, M., Wang, Z. L., & Wood, R. (2018). *The grand challenges of Science Robotics. Science Robotics, 3(14),* eaar7650. https://doi.org/10.1126/scirobotics.aar7650

Yott, J., & Poulin-Dubois, D. (2012). Breaking the rules: Do infants have a true understanding of false belief? *The British Journal of Developmental Psychology, 30(Pt 1),* 156–171. https://doi.org/10.1111/j.2044-835X.2011.02060.x

Zeng, Y., Zhao, Y., Zhang, T., Zhao, D., Zhao, F., & Lu, E. (2020). A Brain-Inspired Model of Theory of Mind. *Frontiers in Neurorobotics, 14,* 60. https://doi.org/10.3389/fnbot.2020.00060

Zhang, C., Bengio, S., Hardt, M., Recht, B., & Vinyals, O. (2021). Understanding deep learning (still) requires rethinking generalization. *Communications of the ACM, 64(3),* 107-115. https://doi.org/10.1145/3446776

Supplemental material

**Supplemental Table 1.** Evidence supporting early infants' Theory of Mind ability through implicit tasks during the first three years of life.

| Study | Infants' age (mo) | Task Type | Paradigm Type (implicit vs explicit) | ToM Measure (behavioural vs neural vs computational) | Main Findings |
|---|---|---|---|---|---|
| *Behavioural* | | | | | |
| *Kovács et al. (2010)* | 7 | Visual object detection task | Implicit | Looking time | Beliefs of an agent (irrelevant to performing the task) modulated infants' looking times, even after the agent had left the scene. |
| *Luo (2011)* | 10 | False-belief task (presence) | Implicit | VOE | Infants associated a preference to the agent with respect to a toy and looked reliably longer when the agent acted in a way that was inconsistent with her preference. |
| *Onishi & Baillargeon (2005)* | 15 | False-belief task (location) | Implicit | VOE | Infants expected the actor to search on the basis of her belief about the toy's location and looked reliably longer when this expectation was violated. |
| *Scott et al. (2010)* | 18 | False-belief task (non-obvious properties) | Implicit | VOE | Infants attributed to an agent a false belief about an object's non-obvious property (rattling noise) and looked longer when they acted in a way that was inconsistent with her false belief. |
| *Scott et al. (2015)* | 17 | False-belief task (identity) | Implicit | VOE | In the deception condition, the infants who saw a deceiving agent replace the rattling test toy with a non-matching silent toy looked reliably longer than those who saw her substitute a matching silent toy. |
| *Senju et al. (2011)* | 18 | False-belief task (location) | Implicit | AL | Anticipatory eye movements revealed that infants who experienced the opaque blindfold expected the actor's action in accord with her having a false belief about the object's location, but infants who experienced the trick blindfold did not. |
| *Song & Baillargeon (2008)* | 14 | False-belief task (identity) | Implicit | VOE | Infants expected the agent to be misled by the tuft's resemblance to the doll's hair and to falsely perceive it as belonging to the doll, as they looked longer when she did not search for the doll in the hair box. |
| *Southgate et al. (2007)* | 25 | False-belief task (location/ presence) | Implicit | AL | Infants correctly anticipated an actor's actions when these actions could be predicted only by attributing a false belief to the actor. |
| *Surian et al. (2007)* | 13 | False-belief task (location) | Implicit | VOE | Infants expected searches for an object to be effective when-- and only when--the agent knew the location of the desired object. |
| *Träuble et al. (2010)* | 15 | False-belief task (location) | Implicit | VOE | Infants accepted visual as well as manual information access as a proper basis for belief induction and looked longer when the agent behaved in a way that was inconsistent with her belief. |

| Yott & Poulin-Dubois (2012) | 18 | False-belief task (location) | Implicit | VOE | After habituating infants to the atypical behavioural rule of looking into box B after seeing the object being places in location A, infants looked significantly longer at the display when the experimenter looked for the toy in the full box (box with the toy) compared to infants who observed the experimenter search in the empty box (box without the toy). |
|---|---|---|---|---|---|
| Moriguchi et al. (2018) | 18 | False-belief task (location) | Implicit | VOE | Infants expected the actor to search in a specific box on the basis of her belief about the toy's location and looked reliably longer when this expectation was violated. |
| *Neural* | | | | | |
| Hyde et al. (2018) | 7 | False-belief task (location) | Implicit | fNIRS | Infants' TPJ activity distinguished between scenarios when another person's belief about the location of the object was false compared with scenarios when the belief was true. |
| Kampis et al. (2015) | 8 | Occlusion events from multiple perspectives | Implicit | EEG | Gamma-band activity was observed (a) when an object was occluded from the infants' perspective, as well as (b) when it was occluded only from the other person, and (c) when subsequently the object disappeared, but the person falsely believed the object to be present. |

| **Southgate & Vernetti (2014)** | 6 | False-belief task (presence) | Implicit | EEG | When an agent had a false belief that a ball was in the box, motor activity in the infant brain (sensorimotor alpha suppression) indicated that infants predicted she would reach for the box. The same was not valid when the agent had a false belief that a ball was not in the box. |
|---|---|---|---|---|---|
| ***Computational + Behavioural*** | | | | | |
| **Hamlin et al. (2013)** | 10 | Social Evaluation Task | Implicit | Looking time + Bayesian modelling | Comparison of computational models involving a mentalistic vs simpler vs non-mentalistic inferences suggested that infants are most likely to engage in mentalistic social evaluation. These results are in concordance with infants' looking times in response to the false-belief task. |

*Abbreviations*- VOE: violation of expectation; AL: anticipatory looking; EEG: electroencephalography; fNIRS: functional near-infrared spectroscopy; TPJ: temporo-parietal junction.

**Supplemental Table 2.** Summary table of developmental psychology experiments describing infants' cognitive ability to predict actions supported by psychological intention recognition theories.

| *Study* | Task | Findings | Supported Computational Theory |
|---|---|---|---|
| *Woodward (1998)* | Prediction of grasping action | After seeing a hand repeatedly grasping one of two objects, infants anticipate that the same object would be grasped again, even when the spatial location of the objects are rearranged | Association |
| *Monroy et al. (2017)* | Prediction of action sequence | Observing actions, but not visual events, influenced toddlers' action choices when associated with an effect | Association |
| *Southgate et al. (2009)* | Prediction of grasping action | Infants display overlapping neural activity during execution and observation of actions, but this activation, rather than being directly induced by the visual input, is driven by infants' understanding of a forthcoming action | Simulation |
| *Skerry et al. (2013)* | Prediction of grasping action | Infants apply a general assumption of efficient action as soon as they have sufficient information (possibly derived from their own action experience) to identify an agent's goal in a given instance | Possibly Simulation for Teleological |
| *Gergely & Csibra (1997)* | Goal attribution to uncompleted action | Infants use the principle of rational action for the interpretation and prediction of goal-directed actions, but also for making productive inferences about unseen aspects of their context | Teleological |
| *Southgate et al. (2008)* | Prediction of goal-directed action | Infants appear to extend goal attribution even to biomechanically impossible actions as long as they are efficient | Teleological |

**Supplemental Table 3.** Summary of psychology experiments describing precursor computational models of human Theory of Mind ability.

| Study | Precursor of Theory of Mind | Proposal | Experiments |
|---|---|---|---|
| *Keysers & Gazzola (2007)* | Simulation as precursor of Theory of Mind | Based on neural evidence: brain areas associated with both accounts represent simulation (even though at different levels) | N/A |
| *Gergely & Csibra (1997)* | Teleological as precursor of Theory of Mind | Continuum between teleological constructs (i.e., action, goal-state and situational constraints) and mentalistic ones (i.e., intentions, desires and beliefs), with the latter supposing same computations and constructs of the former but representing more sophisticated, abstract constructs | 12-month-old infants had to infer a goal state to rationalise the incomplete action, whose end state was occluded from them, as an efficient 'chasing' action |
| *Baker et al. (2017)* | Teleological as precursor of Theory of Mind | Bayesian computational model for ToM based on the teleological principle: adults were suggested to follow this model to infer the mental states behind an agent's behaviour using priors | 'Food-trucks' scenario, using animated two-dimensional displays of an agent navigating through simple grid-worlds. Observers had to infer agents' mental states according to their trajectory |
| *Hamlin et al. (2013)* | Teleological as precursor of Theory of Mind | Bayesian computational model for ToM based on the teleological principle: infants were suggested to follow this model to infer the mental states behind an agent's behaviour | Social evaluation task assessing infants' judgement of intentions (helpful and harmful) of observed agents during interaction |

**Supplemental Table 4.** Summary of relevant robotic or computational implementations in the literature addressing inference vs learning of others' mental states vs learning of others' beliefs.

| Study | Inference of others' mental states | Learning of others' mental states BUT beliefs | Explicit Learning of others' beliefs |
|---|---|---|---|
| *Devin & Alami (2016)* | ✓ | X | X |
| *Görür et al. (2017)* | ✓ | X | X |
| *Hiatt et al. (2011)* | ✓ | X | X |
| *Demiris & Khadhouri (2006)* | ✓ | X | X |
| *Winfield (2018)* | ✓ | X | X |
| *Baker et al. (2017)* | ✓ | X | X |
| *Hamlin et al. (2013)* | ✓ | X | X |
| *Patacchiola & Cangelosi (2016)* | ✓ | X | X |
| *Asakura & Inui (2016)* | ✓ | X | X |
| *Ramirez & Geffner (2011)* | ✓ | X | X |
| *Kominis & Geffner (2015, 2017)* | ✓ | X | X |
| *Zeng et al. (2020)* | ✓ | X | X |
| *Raileanu et al. (2018)* | ✓ | ✓ | X |
| *He et al. (2016)* | ✓ | ✓ | X |
| *Chen et al. (2021)* | ✓ | ✓ | X |
| *Nguyen & Gonzalez (2022)* | ✓ | ✓ | ✓ |
| *Rabinowitz et al. (2018)* | ✓ | ✓ | ✓ |